\documentclass[sigconf,nonacm]{acmart}
\AtBeginDocument{%
  }

\renewcommand\footnotetextcopyrightpermission[1]{}
\makeatletter
\def\@copyrightpermission{}
\makeatother

\makeatletter
\fancypagestyle{standardpagestyle}{%
  \fancyhf{}%
}
\fancypagestyle{firstpagestyle}{%
  \fancyhf{}%
}
\makeatother

\usepackage{dsfont}
\usepackage{balance}
\usepackage{tabularx}
\usepackage{multirow}

\begin{document}

\title[]{Stable Diffusion-Based Approach for Human De-Occlusion}

\author{Seung Young Noh}
\email{kelvinnoh@kw.ac.kr}
\affiliation{%
  \institution{Kwangwoon University}
  \city{Seoul}
  \country{South Korea}
}

\author{Ju Yong Chang}
\email{jychang@kw.ac.kr}
\affiliation{%
  \institution{Kwangwoon University}
  \city{Seoul}
  \country{South Korea}
}

\renewcommand{\shortauthors}{}

\begin{abstract}
Humans can infer the missing parts of an occluded object by leveraging prior knowledge and visible cues. However, enabling deep learning models to accurately predict such occluded regions remains a challenging task. De-occlusion addresses this problem by reconstructing both the mask and RGB appearance. In this work, we focus on human de-occlusion, specifically targeting the recovery of occluded body structures and appearances. Our approach decomposes the task into two stages: mask completion and RGB completion. The first stage leverages a diffusion-based human body prior to provide a comprehensive representation of body structure, combined with occluded joint heatmaps that offer explicit spatial cues about missing regions. The reconstructed amodal mask then serves as a conditioning input for the second stage, guiding the model on which areas require RGB reconstruction. To further enhance RGB generation, we incorporate human-specific textual features derived using a visual question answering (VQA) model and encoded via a CLIP encoder. RGB completion is performed using Stable Diffusion, with decoder fine-tuning applied to mitigate pixel-level degradation in visible regions---a known limitation of prior diffusion-based de-occlusion methods caused by latent space transformations. Our method effectively reconstructs human appearances even under severe occlusions and consistently outperforms existing methods in both mask and RGB completion. Moreover, the de-occluded images generated by our approach can improve the performance of downstream human-centric tasks, such as 2D pose estimation and 3D human reconstruction. The code will be made publicly available. 
\end{abstract}





\maketitle

\thispagestyle{firstpagestyle}
\pagestyle{standardpagestyle}

\section{Introduction}

When capturing the 3D world with a camera, the scene is projected onto a 2D image, often leading to occlusion. Occlusion occurs when an object is partially or fully blocked by another, resulting in the loss of visual information. This presents significant challenges for deep learning models in accurately understanding occluded objects. In contrast, humans can naturally infer missing regions by leveraging visible cues and contextual information. In particular, when humans are occluded, it is often possible to estimate the shape and color of hidden body parts based on visible regions, body pose, and overall appearance.

\begin{figure}
    \centering
    \includegraphics[width=0.9\linewidth]{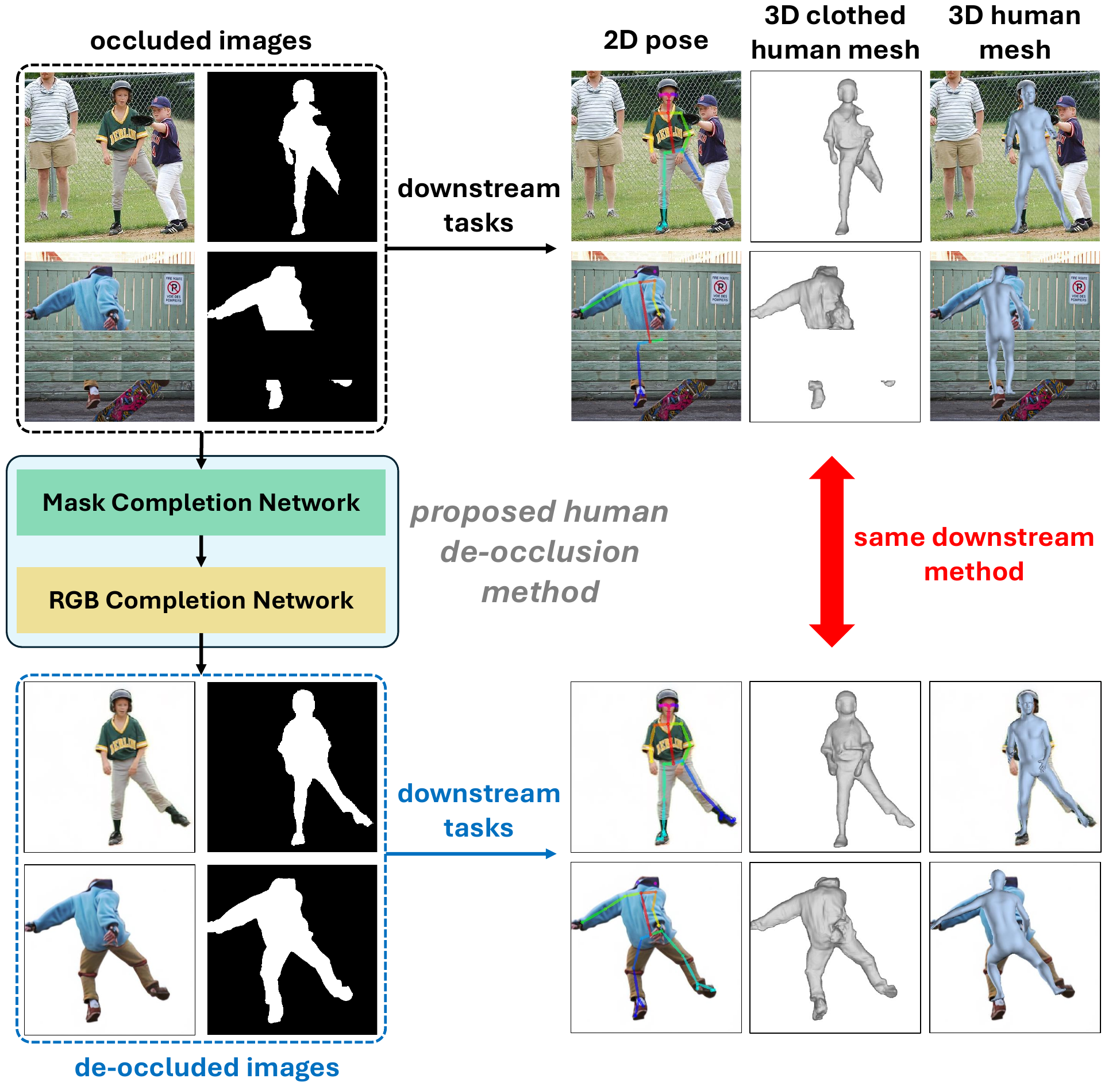}
    \vspace{-3mm}
    \caption{Our human de-occlusion method reconstructs de-occluded images from occluded inputs by using both a mask completion network and an RGB completion network. The resulting images can be utilized as improved inputs for downstream tasks such as 2D pose estimation~\cite{8765346}, 3D clothed human reconstruction~\cite{saito2019pifu}, and 3D human mesh recovery~\cite{kanazawaHMR18}, leading to better performance across these tasks.}
    \label{fig:1}
    \vspace{-4mm}
\end{figure}

Human de-occlusion aims to reconstruct both the mask and RGB appearance of occluded body regions in images. By recovering missing information, it enables the generation of images that closely resemble the original, unoccluded form. As shown in Fig.~\ref{fig:1}, such reconstructions support various downstream tasks, including 2D pose estimation, 3D clothed human reconstruction, and 3D human mesh recovery.

Due to its practical advantages, de-occlusion has been an active research topic, often tackled by extending the outputs from existing vision tasks. For example,~\cite{kar2015amodal,hsieh2023tao} proposed amodal bounding box completion to account for occluded regions, while~\cite{li2016amodal,zheng2021visiting,ling2020variational,follmann2019learning,zhu2017semantic,qi2019amodal,dhamo2019object,li2023muva,hu2019sail,reddy2022walt,vuong2024walt3d,ke2021deep} extended instance segmentation to amodal instance segmentation, which incorporates occluded areas into segmentation outputs. Additionally,~\cite{zhan2024amodal,nguyen2021weakly} introduced amodal completion networks to infer the amodal mask from a modal (visible) mask. While these methods are effective in recovering occlusion masks, they remain limited in reconstructing RGB appearance.

De-occlusion, in a broader sense, refers to the task of reconstructing both the mask and RGB appearance of occluded regions. For example,~\cite{zhan2020self} proposed a method that first predicts an amodal mask based on occlusion order and then restores the RGB values. Due to the challenges of de-occlusion across diverse object types, several studies have focused on category-specific approaches (e.g., furniture~\cite{ehsani2018segan}, hands~\cite{meng20223d}, humans~\cite{zhou2021human}, vehicles~\cite{yan2019visualizing,ling2020variational}, and pizza~\cite{papadopoulos2019make}) to improve performance. However, these methods often rely on limited datasets, which restrict their generalizability, especially in zero-shot settings. To address this limitation, recent works~\cite{liu2024object,ozguroglu2024pix2gestalt,xu2024amodal} have utilized diffusion models (DMs)~\cite{ho2020denoising,rombach2022high}, leveraging rich image priors to enhance zero-shot de-occlusion performance. While effective across various categories, these methods struggle with human de-occlusion due to the lack of structured human body priors. Unlike existing methods, humans intuitively infer missing body parts using physical and structural constraints---capabilities that diffusion priors alone have difficulty replicating.

Furthermore, RGB completion methods commonly utilize either U-Net~\cite{ronneberger2015u} or Stable Diffusion~\cite{rombach2022high} to reconstruct occluded pixel regions. While U-Net-based approaches can recover the coarse appearance of occluded regions, they often produce overly smoothed outputs lacking high-frequency details. To address this, recent works~\cite{liu2024object,ozguroglu2024pix2gestalt,xu2024amodal} have adopted Stable Diffusion for its ability to leverage rich image priors. However, due to the inherent pixel-to-latent transformation, Stable Diffusion often suffers from information loss, particularly in preserving fine-grained visible RGB details. In addition, Stable Diffusion-based methods typically condition on text features extracted from the input RGB image using a pre-trained CLIP encoder~\cite{radford2021learning}, which may inadvertently encode both the foreground human and irrelevant background context. They also rely on generic textual prompts (e.g., "a photo of a <class name>", "a <class name>") to describe the scene. As an alternative, incorporating more refined and human-specific text features could enhance de-occlusion performance.

To address these challenges, we propose a two-stage human de-occlusion framework comprising mask completion and RGB completion, inspired by prior two-stage approaches~\cite{zhan2020self,ehsani2018segan,xu2024amodal,meng20223d,zhou2021human,yan2019visualizing}, but with specialized modules for each stage. In the mask completion stage, rather than relying on object-centric priors from Stable Diffusion, we use a \emph{diffusion-based human body prior} that provides structural guidance for occluded human bodies. This prior is derived from 2D joint coordinates and heatmaps using established techniques from human mesh recovery~\cite{zhu2024dpmesh}. To explicitly convey information about the occluded regions, we also introduce an \emph{occluded joint heatmap}, inspired by eraser masks commonly used in de-occlusion~\cite{zhan2020self,liu2024object,xu2024amodal}. This heatmap enhances both the quantitative and qualitative accuracy of mask completion.

In the RGB completion stage, we use the amodal mask predicted in the previous stage as a condition to guide the pixel-level reconstruction. This explicit localization enables the model to focus its generative capacity on the occluded regions, improving reconstruction fidelity.

Prior methods typically employ a frozen CLIP encoder to extract text features from either the input image~\cite{ozguroglu2024pix2gestalt,xu2024amodal} or simple textual prompts~\cite{zhao2023unleashing,liu2024object}, which may lack fine-grained, human-specific information. To overcome this limitation, we generate a human-centric sentence using a visual question answering (VQA) model~\cite{li2022blip}, offering a more focused and context-aware textual representation. The generated sentence is then encoded by a CLIP encoder to produce \emph{human-specific text features} containing rich semantic cues about human appearance and attributes. These features help the model infer invisible RGB values more accurately by emphasizing meaningful visual cues.

To further mitigate the information loss caused by the latent space transformation in Stable Diffusion, we employ \emph{decoder fine-tuning}~\cite{avrahami2023blended}, which enhances appearance-level fidelity. Our experimental results demonstrate that the proposed method improves RGB restoration in occluded regions and better preserves visible region details that are often lost in existing approaches due to latent space conversion.

The key contributions of this work are:
\begin{itemize}
    \item We propose to use a diffusion-based human body prior and an occluded joint heatmap to provide structural guidance for accurate amodal mask prediction under human occlusion.
    \item We use the amodal mask as a conditioning input for RGB completion to localize the reconstruction regions, and apply decoder fine-tuning to mitigate information loss caused by the latent space transformation in Stable Diffusion.
    \item We extract human-specific text features by generating human-centric sentences via a VQA model, enhancing the semantic guidance provided to the CLIP encoder and improving RGB reconstruction quality. 
\end{itemize}

\section{Related Works}

\subsection{De-occlusion}

Instance segmentation~\cite{he2017maskrcnn,liu2018panet,bolya2019yolact,wang2019solo,wang2020solov2,tian2020condinst} is a type of image segmentation that aims to extract segmentation masks based solely on visible pixels in a given RGB image. ~\cite{li2016amodal} redefined the existing instance segmentation task as modal instance segmentation and introduced a new task, amodal instance segmentation, which incorporates both visible and occluded pixels into the segmentation process. Amodal completion is closely related to amodal instance segmentation~\cite{li2016amodal,follmann2019learning,liu2019layered_amodal,back2021unseen_amodal,ehsani2018amodal_seg}, with the key difference being that it explicitly takes a visible mask (modal mask) as input~\cite{zhan2020self,zhan2024amodal}. In other words, amodal instance segmentation predicts the amodal mask directly from an RGB image, whereas amodal completion reconstructs the amodal mask given both the RGB image and the modal mask. Amodal appearance completion further extends amodal completion by not only predicting the amodal mask but also recovering the RGB appearance of occluded regions~\cite{liu2024object}, a process sometimes referred to as de-occlusion. Since different papers use the terms amodal completion, amodal appearance completion, and de-occlusion interchangeably (e.g., using amodal completion to refer to amodal appearance completion), we consistently use de-occlusion throughout this paper to avoid confusion.

SSSD~\cite{zhan2020self} performs de-occlusion using PCNet-M and PCNet-C, both based on U-Net. PCNet-M reconstructs the mask based on the predicted occlusion order graph, while PCNet-C reconstructs RGB values. They also proposed a self-supervised method that can be trained without ground truth (GT); however, it has the limitation of requiring an eraser mask that induces occlusion as an input to the network. VINV~\cite{zheng2021visiting} utilizes an occlusion order graph to group objects of the same graph depth into a single layer. De-occlusion is then performed sequentially, starting with objects at the lowest graph depths. They further propose an iterative approach in which the recovered RGB image at each stage serves as the input for the next stage of de-occlusion.
The diversity of objects increases the difficulty of the de-occlusion task, leading some works to restrict de-occlusion to specific categories (e.g., furniture~\cite{ehsani2018segan}, vehicles~\cite{yan2019visualizing,ling2020variational}). In~\cite{zhou2021human}, the de-occlusion target is limited to humans, and a U-Net-based model is proposed to learn human body information using pseudo-GT human parsing images. However, this method is impractical in real-world scenarios as it relies on the GT amodal mask and GT human parsing image instead of using the amodal mask obtained through mask completion during RGB completion. Recently, with the widespread use of DMs across various fields, research has explored two approaches for applying them to de-occlusion: fine-tuning a pre-trained diffusion network for the task~\cite{ozguroglu2024pix2gestalt,liu2024object}, or iteratively applying a pre-trained diffusion network without fine-tuning until the target object is fully restored~\cite{xu2024amodal}.

De-occlusion can reconstruct an object to its fully-visible state, and several works have reported performance improvements when applying it to downstream tasks. For example,~\cite{meng20223d} improved 3D hand pose estimation by de-occluding hand images, while~\cite{ozguroglu2024pix2gestalt} reported improvements in image segmentation, object recognition, and 3D reconstruction.

\vspace{-2mm}
\subsection{Image Inpainting}

Image inpainting aims to generate realistic results by filling in missing regions or modifying specific objects in an input image. Early research explored approaches based on the generative adversarial network~\cite{goodfellow2014gan}, leading to several notable works~\cite{pathak2016context_inpaint,yu2018generative_inpaint,yang2017high_inpaint,yu2019free_inpaint,nazeri2019edgeconnect_inpaint}. More recently, DMs have been used for inpainting, demonstrating their ability to generate visually coherent and high-quality results~\cite{lugmayr2022repaint,suvorov2021lama,li2022mat,xie2023smartbrush,ju2024brushnet}. Image inpainting and de-occlusion both recover missing areas in RGB images, but they differ in their input methods and objectives. Image inpainting takes an explicitly defined region as input, whereas de-occlusion detects occluded regions autonomously through amodal completion. The RGB reconstruction process also differs between the two tasks. While image inpainting primarily focuses on generating colors that blend seamlessly with the surrounding regions, de-occlusion is further constrained to recovering the RGB values of the physically occluded parts of the target object. Due to this distinction, even when GT masks for the occluded regions are available, image inpainting methods often struggle with de-occlusion, despite recent advances in performance~\cite{liu2024object}.

\subsection{Diffusion Models}

Recently, the DM~\cite{ho2020denoising} has shown strong performance in image generation tasks. Stable Diffusion~\cite{rombach2022high}, which employs a variational autoencoder (VAE)~\cite{kingma2013auto} to perform the reverse diffusion process~\cite{sohl2015deep} in a latent space instead of pixel space for improved efficiency, has been widely adopted. Trained on the large-scale image-text dataset LAION-5B~\cite{schuhmann2022laion}, it has shown a strong ability to generate diverse and high-quality images, making it applicable in various domains. ~\cite{zhang2023adding} proposed ControlNet, which enables controlled image generation based on conditions such as human pose, depth maps, or normal maps. Pre-trained DMs capture rich prior knowledge about various objects, and when combined with the ability of ControlNet to generate conditional images, they have been applied not only to image-related tasks such as image generation, inpainting, and novel view synthesis, but also to 3D-related tasks. The Visual Perception with Diffusion framework, which employs a pre-trained DM as a backbone for visual perception tasks, was proposed in ~\cite{zhao2023unleashing}. This approach leverages the prior knowledge embedded in DMs as features and has been validated in various tasks, including text-image alignment~\cite{kondapaneni2024text}, human mesh recovery~\cite{zhu2024dpmesh}, and depth estimation~\cite{patni2024ecodepth,zhao2023unleashing}.

\subsection{Occlusion Handling in Downstream Tasks}

Our method produces a de-occluded image from occluded input, which can be utilized in various downstream tasks. Among them, human mesh recovery is a representative example, in which occlusion-robust methods~\cite{cheng20203d,kocabas2021pare,huang2022object,moon20223d,zhang20233d,gwon2025eigenpose} have been proposed to infer pose and shape directly from occluded inputs. While these methods perform well under moderate occlusion, they often struggle with severe occlusions. In such cases, our de-occluded images can improve performance not only for models that do not explicitly handle occlusion but also for those incorporating occlusion-aware mechanisms. Further qualitative results and analysis are provided in Section C.8 of the supplementary material.

\section{Proposed Method}

\subsection{Preliminaries: Diffusion Models}

DM is a generative model that learns a reverse process to progressively remove noise, thereby recovering the data distribution from noise. In the denoising U-Net, cross-attention is computed between text feature extracted from a pre-trained CLIP~\cite{radford2021learning} encoder and the image feature, guiding the denoising process.

Recently, ControlNet~\cite{zhang2023adding} has been utilized by incorporating a trainable copy of the Stable Diffusion encoder block, enabling conditional image generation. To train the ControlNet and the denoising U-Net, the input image $I$ is encoded into a noisy latent representation $z_t$ using a pre-trained VAE encoder $\mathcal{E}$, formulated as:
\begin{equation}
\label{eq1}
    z_0 = \mathcal{E}(I),\;
    z_t = \sqrt{\bar{\alpha}_t} z_0 + \sqrt{1 - \bar{\alpha}_t} \epsilon,
\end{equation}
where $t$ denotes the timestep, $\bar{\alpha}_t = \prod_{s=1}^{t} \alpha_s$, and $\epsilon \sim \mathcal{N}(0, I)$.

The training objective for the denoising U-Net $\epsilon_\theta$ and the ControlNet $\Psi_\text{c}$ with the conditioning input $\mathcal{C}$ is defined as:
\begin{equation}
\label{eq2}
    \mathcal{L} = \mathbb{E}_{z_0, \epsilon, t, \mathcal{C}} \left[ \|\epsilon - \epsilon_\theta (z_t, t, \mathcal{C}) \|_2^2 \right],
\end{equation}
where $\mathcal{C}$ typically consists of a text embedding $f_\text{t}$ extracted from a frozen CLIP encoder and a control embedding $f_\text{c}$ derived from $\Psi_\text{c}$, forming $\mathcal{C} = \{f_\text{t}, f_\text{c}\}$. $f_\text{t}$ can be extracted from the input image $I$, and $f_\text{c}$ can be obtained from various inputs such as a canny edge map, human pose, etc. The control feature $f_\text{c}$ is injected into the decoder block of the denoising U-Net $\epsilon_\theta$ to induce the condition to be present in the output image.

\subsection{Overview}

We propose a method for reconstructing a de-occluded amodal mask and a complete RGB image from a human modal mask and an occluded RGB image. Our approach follows a two-stage framework consisting of mask completion and RGB completion, where de-occlusion is performed progressively. By adopting this two-stage framework, the amodal mask predicted in the first stage serves as a condition for RGB completion, allowing each network to specialize in detecting occluded regions and reconstructing their corresponding RGB values. The design choice and supporting experiments for the two-stage framework are described in Section~\ref{subsec:ab}.

First, the mask completion network $\Phi_\text{mask}$ takes the modal mask $M_\text{m} \in \mathbb{R}^{H \times W}$ and the occluded RGB image $I_\text{o} \in \mathbb{R}^{H \times W \times 3}$ as inputs, generating the amodal mask $M_\text{a} \in \mathbb{R}^{H \times W}$. Next, the RGB completion network $\Phi_\text{rgb}$ utilizes the modal mask $M_\text{m}$, the reconstructed amodal mask $M_\text{a}$, and the occluded image $I_\text{o}$ to restore the de-occluded RGB image $I_\text{do} \in \mathbb{R}^{H \times W \times 3}$. This process can be formulated as:
\begin{equation}
\label{eq3}
    M_\text{a} = \Phi_\text{mask} (M_\text{m}, I_\text{o}),\;
    I_\text{do} = \Phi_\text{rgb} (M_\text{m}, M_\text{a}, I_\text{o}).
\end{equation}
Mask completion and RGB completion networks are described in detail in Sections~\ref{subsec:mask} and~\ref{subsec:rgb}, respectively.

\begin{figure}
    \centering
    \includegraphics[width=0.9\linewidth]{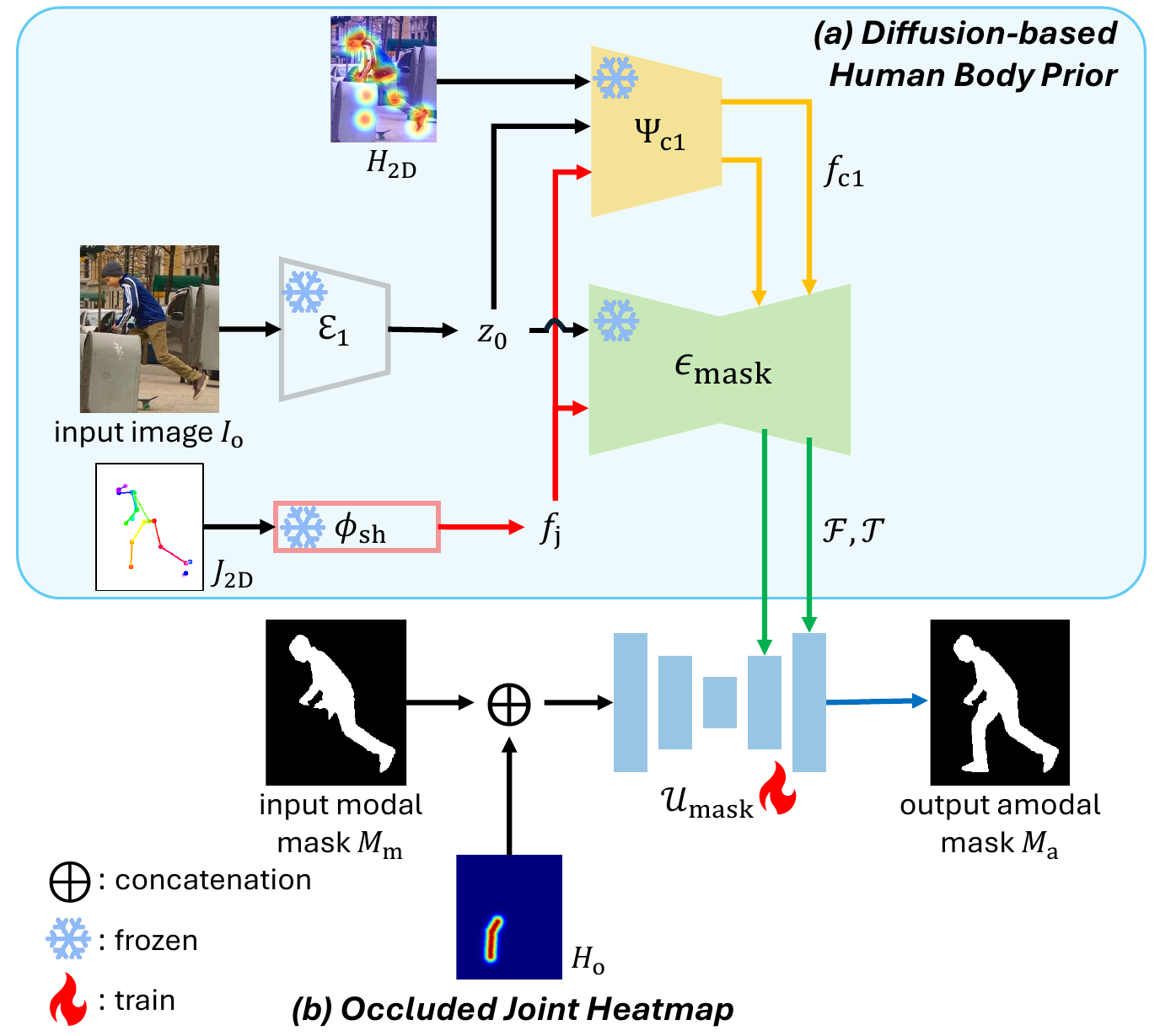}
    \vspace{-4mm}
    \caption{Overview of mask completion network.}
    \label{fig:2}
    \vspace{-5mm}
\end{figure}

\subsection{Mask Completion Network}
\label{subsec:mask}

The objective of the mask completion network $\Phi_\text{mask}$ is to reconstruct the amodal mask $M_\text{a}$ by leveraging human body priors, given an occluded human RGB image $I_\text{o}$ and a modal mask $M_\text{m}$. To this end, we extract a diffusion-based human body prior using the DM from~\cite{ozguroglu2024pix2gestalt}, which is fine-tuned with Stable Diffusion for the human mesh recovery task. In addition, we generate occluded joint heatmaps that provide more explicit cues about the missing body parts and incorporate them into the mask completion process. Using this information, $\Phi_\text{mask}$ predicts an amodal mask that recovers the occluded regions, which is illustrated in Fig.~\ref{fig:2}.

\noindent\textbf{Diffusion-based human body prior.}  
Following~\cite{zhan2024amodal,zhao2023unleashing,zhu2024dpmesh,kondapaneni2024text,patni2024ecodepth}, the diffusion-based human body prior is extracted using a pre-trained denoising U-Net as the image backbone, which produces multi-scale feature maps and cross-attention maps. Given the occluded RGB image $I_\text{o}$, an off-the-shelf 2D pose estimator $\phi_\text{pose}$~\cite{8765346} extracts $k$ 2D joint coordinates $J_\text{2D} \in \mathbb{R}^{k \times 2}$. The extracted $J_\text{2D}$ is then used to generate a heatmap $H_\text{2D} \in \mathbb{R}^{k \times h \times w}$, formulated as:
\begin{equation}
\label{eq4}
    J_\text{2D} = \phi_\text{pose}(I_\text{o}),\;
    H_\text{2D} = Gaussian(J_\text{2D}, \sigma_1),
\end{equation}
where $Gaussian(\cdot)$ generates Gaussian heatmaps centered at the given joint coordinates using a 2D Gaussian kernel, and $\sigma_1$ controls the spatial spread of the Gaussian kernel.

In conventional denoising U-Net architectures, cross-attention maps are computed between text feature extracted from a CLIP encoder and image feature. However, following~\cite{zhu2024dpmesh}, we replace the text feature with joint feature $f_\text{j} \in \mathbb{R}^{k \times 768}$, extracted from visible 2D joints using a shallow multi-layer perceptron $\phi_\text{sh}$. Additionally, the input image $I_\text{o}$ is transformed into a latent representation $z_0 \in \mathbb{R}^{4 \times h \times w}$ via a pre-trained VAE encoder $\mathcal{E}_1$, formulated as:
\begin{equation}
\label{eq5}
    f_\text{j} = \phi_\text{sh}(J_\text{2D}),\;
    z_0 = \mathcal{E}_1(I_\text{o}).
\end{equation}

The 2D heatmap $H_\text{2D}$ from Eq.~\eqref{eq4} is provided as an input to the ControlNet $\Psi_\text{c1}$, along with the latent representation $z_0$ and the joint feature $f_\text{j}$. The ControlNet $\Psi_\text{c1}$ processes these features to extract a control feature $f_\text{c1}$, formulated as:
\begin{equation}
\label{eq6}
f_\text{c1} = \Psi_\text{c1}(H_\text{2D}, z_0, f_\text{j}),
\end{equation}
where $f_\text{c1}$ is subsequently injected into the decoder blocks of the denoising U-Net $\epsilon_\text{mask}$, guiding the network to produce a more meaningful diffusion-based human body prior.

The joint feature $f_\text{j}$ from Eq.~\eqref{eq5} and the control feature $f_\text{c1}$ from Eq.~\eqref{eq6} are combined to form the conditioning input, defined as $\mathcal{C}_1 = \{ f_\text{j}, f_\text{c1} \}$. The denoising U-Net $\epsilon_\text{mask}$ takes the latent representation $z_0$ of the input image, along with the conditioning input $\mathcal{C}_1$ and the timestep $t$, to produce multi-scale feature maps $\mathcal{F}$ and cross-attention maps $\mathcal{T}$, formulated as:
\begin{equation}
\label{eq7}
    \mathcal{F}, \mathcal{T} = \epsilon_\text{mask}(z_0, t, \mathcal{C}_1),
\end{equation}
where $t$ is set to $0$ to prevent noise from being added to the latent representation, following~\cite{zhao2023unleashing,kondapaneni2024text}. The spatial resolution of each feature map $\mathcal{F}_m$ is given by $H_m = W_m = 2^{(m+1)}$, for $m = 1, 2, 3, 4$. Similarly, each cross-attention map $\mathcal{T}_n$ has spatial dimensions $H_n = W_n = 2^{(n+2)}$, for $n = 1, 2$.

\noindent\textbf{Occluded joint heatmap.}
In addition to the diffusion-based human body prior, which provides comprehensive information about the human body, we construct an occluded joint heatmap based on 2D joint coordinates $J_\text{2D}$ to explicitly inform the network about occluded regions. First, we exclude six foot joints (four toes and two heels) to obtain $J_\text{body} \in \mathbb{R}^{(k-6) \times 2}$, as these joints are often occluded or have low confidence due to detection errors.

Traditional 2D joint-based human pose representations use a limited number of joints for efficient model training and inference. However, these joints are widely spaced, forming a sparse representation that struggles to capture the continuous structure of the human body. Inspired by subdivision techniques in 3D Gaussian representations and meshes~\cite{wen2025life,wang2023octformer} for improved resolution, we apply interpolation between adjacent joints to generate subdivided joint coordinates $J_\text{sub} \in \mathbb{R}^{s \times 2}$. By concatenating $J_\text{sub}$ with the original 2D joint set $J_\text{body}$, we obtain a denser joint representation $J_\text{dense} \in \mathbb{R}^{(k-6+s) \times 2}$. Among all joints in $J_\text{dense}$, we select only those located in occluded regions, as indicated by the modal mask $M_\text{m}$, forming the occluded joint set $J_\text{o} \in \mathbb{R}^{o \times 2}$:
\begin{equation}
\label{eq8}
    J_\text{o} = \{ j_i \in J_\text{dense} \mid \mathds{1}(M_\text{m} (j_i) = 0) = 1 \},
\end{equation}
where $j_i$ represents each joint in $J_\text{dense}$, and $\mathds{1}(\cdot)$ is an indicator function that returns 1 if the condition is met and 0 otherwise.

$J_\text{o}$ consists of joints that are likely present but labeled as zero in the modal mask, indicating a probability of occlusion. To incorporate this information into the mask completion network, we construct an occluded joint heatmap $H_\text{o} \in \mathbb{R}^{H \times W}$, defined as:
\begin{equation}
\label{eq9}
    H_\text{o} = \max \left( {Gaussian}(J_\text{o}, \sigma_2) \right),
\end{equation}
where $Gaussian(J_\text{o}, \sigma_2) \in \mathbb{R}^{o \times H \times W}$ denotes a stack of 2D Gaussian heatmaps, each centered at a joint in $J_\text{o}$, as in Eq.~\eqref{eq4}. The $\max$ operation is applied pixel-wise along the joint dimension to produce a single-channel heatmap, and $\sigma_2$ determines the spatial extent of the Gaussian kernel.

\noindent\textbf{Mask completion.}  
The amodal mask $M_\text{a}$ is reconstructed by the U-Net $\mathcal{U}_\text{mask}$, which takes as input the modal mask $M_\text{m}$, the occluded joint heatmap $H_\text{o}$, and the multi-scale feature maps $\mathcal{F}$ along with the cross-attention maps $\mathcal{T}$ obtained from Eq.~\eqref{eq7}. This process is formulated as:
\begin{equation}
\label{eq10}
    M_\text{a} = \mathcal{U}_\text{mask}( M_\text{m} \oplus H_\text{o}, \mathcal{F}, \mathcal{T}),
\end{equation}
where $\oplus$ denotes concatenation operation.

To train the $\mathcal{U}_\text{mask}$, we minimize the binary cross-entropy loss between the predicted amodal mask $M_\text{a}$ and the ground-truth amodal mask $M_\text{a}^*$, which is computed as:
\begin{equation}
\label{eq11}
    \mathcal{L}_{\text{mask}} = - \frac{1}{N} \sum_{i=1}^{N} \big( p^*_i \log p_i + (1 - p^*_i) \log (1 - p_i) \big),
\end{equation}
where $p_i = M_\text{a}(i)$ and $p^*_i = M_\text{a}^*(i)$ denote the predicted and ground truth amodal mask values at pixel $i$, respectively, and $N$ is the total number of pixels.

\begin{figure}
    \centering
    \includegraphics[width=0.9\linewidth]{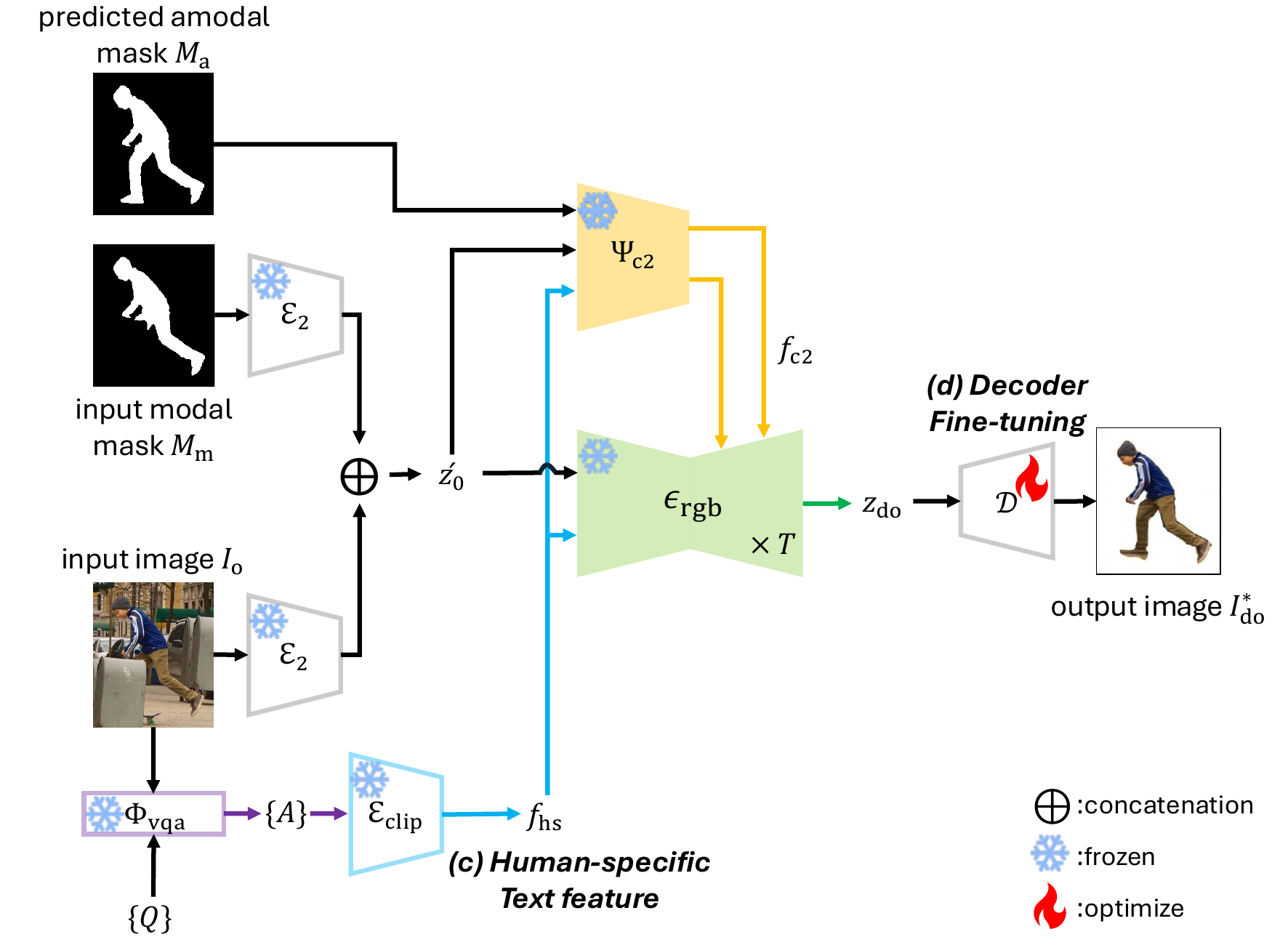}
    \vspace{-4mm}
    \caption{Overview of RGB completion network.}
    \label{fig:3}
    \vspace{-4mm}
\end{figure}

\subsection{RGB Completion Network}
\label{subsec:rgb}

The objective of the RGB completion network $\Phi_\text{rgb}$ is to reconstruct the de-occluded image $I_\text{do}$ from the occluded image $I_\text{o}$, the modal mask $M_\text{m}$, and the amodal mask $M_\text{a}$ predicted by the mask completion network $\Phi_\text{mask}$. To this end, we adopt the approach proposed in~\cite{ozguroglu2024pix2gestalt}, which fine-tunes Stable Diffusion for the de-occlusion task. In addition, we produce human-centric descriptions using VQA model~\cite{li2022blip}, instead of directly extracting text features from the input image via CLIP~\cite{radford2021learning}, which may include background information that is not directly related to the human subject. These descriptions are fed into the CLIP encoder to extract semantically aligned features for guiding the RGB completion. While these human-specific text features provide meaningful guidance, the use of latent representations may result in the loss of fine-grained details, especially in visible areas. To address this issue, we fine-tune the decoder to improve pixel-level fidelity. The overall process of the RGB completion network is shown in Fig.~\ref{fig:3}.

\noindent\textbf{Human-specific text feature.}
To obtain human-specific text features, we utilize a VQA model $\Phi_\text{vqa}$~\cite{li2022blip} to generate descriptive sentences about the person in the input image $I_\text{o}$. Following~\cite{huang2024tech}, we employ a set of fine-grained VQA questions $\{Q\}$ designed to capture detailed information about garments (e.g., pants, skirt, shoes, etc.) in terms of color and style, as well as human attributes (e.g., pose, hair). These questions $\{Q\}$ are fed into the $\Phi_\text{vqa}$ along with $I_\text{o}$, yielding responses $\{A\}$ that describe the person's garments and appearance. Finally, we encode $\{A\}$ using a CLIP encoder $\mathcal{E}_\text{clip}$ to extract the human-specific text feature $f_\text{hs}$. This process is formulated as:
\begin{equation}
\label{eq12}
    \{A\}=\Phi_\text{vqa}(I_\text{o},\{Q\}),\;
    f_\text{hs}=\mathcal{E}_\text{clip}(\{A\}).
\end{equation}

\noindent\textbf{RGB completion.}
In the RGB completion stage, we use the amodal mask $M_\text{a}$, obtained from the mask completion stage, as a condition to provide explicit information about the occluded regions and guide the reconstruction process. Specifically, we extract the control feature $f_\text{c2}$ by feeding $M_\text{a}$ into ControlNet $\Psi_\text{c2}$, in a manner similar to the mask completion formulation in Eq.~\eqref{eq6}, as follows:
\begin{equation}
\label{eq13}
    z_0^\prime=\mathcal{E}_2(I_\text{o}) \oplus \mathcal{E}_2(M_\text{m}),\;
    f_\text{c2} = \Psi_\text{c2}(M_\text{a},z_0^\prime,f_\text{hs}),
\end{equation}
where $\mathcal{E}_2$ is the VAE encoder used in the RGB completion stage, and $z_0^\prime$ is formed by concatenating the latent representations of the occluded image and the modal mask, $\mathcal{E}_2(I_\text{o})$ and $\mathcal{E}_2(M_\text{m})$.

The human-specific text feature $f_\text{hs}$ from Eq.~\eqref{eq12} and the control feature $f_\text{c2}$ from Eq.~\eqref{eq13} are defined as the conditioning input $\mathcal{C}_2 = \{f_\text{hs}, f_\text{c2}\}$. The denoising U-Net $\epsilon_\text{rgb}$ takes the latent representation $z_0^\prime$, the timestep $t$, and the conditioning input $\mathcal{C}_2$ as input, and outputs the predicted noise $\epsilon_t$ corresponding to the current timestep, formulated as
\begin{equation}
\label{eq14}
    \epsilon_t = \epsilon_\text{rgb}(z_0^\prime, z_t, t, \mathcal{C}_2),
\end{equation}
where $0 \leq t \leq T$, and $z_t$ denotes the latent representation at step $t$ in the reverse denoising process, with initialization $z_T \sim \mathcal{N}(0, I)$.

Using the predicted noise $\epsilon_t$ from Eq.~\eqref{eq14}, we first compute the predicted de-occluded latent $\hat{z}_\text{do}$ in a deterministic manner. Then, the latent representation at the next step in the reverse process, $z_{t-1}$, is obtained from $\hat{z}_\text{do}$ and $\epsilon_t$ as follows:
\begin{equation}
\label{eq15}
    \hat{z}_\text{do} = \frac{z_t - \sqrt{1 - \bar{\alpha}_t} \epsilon_t}{\sqrt{\bar{\alpha}_t}},\;
    z_{t-1} = \sqrt{\bar{\alpha}_{t-1}} \hat{z}_\text{do} + \sqrt{1 - \bar{\alpha}_{t-1}} \epsilon_t,
\end{equation}
where $\bar{\alpha}_t$ is the cumulative noise schedule at timestep $t$. This process is iteratively applied by feeding $z_{t-1}$ back into Eq.~\eqref{eq14} for the next denoising step. Repeating this reverse process from $t = T$ down to $t = 1$ yields the final de-occluded latent representation $z_0 = z_\text{do}$.

\noindent\textbf{Deocder fine-tuning.}
However, reconstructing $I_\text{do}$ by decoding the predicted latent $z_\text{do}$ using the VAE decoder $\mathcal{D}$ often results in a loss of pixel-level detail in the non-occluded regions. To address this issue, instead of retraining the noise prediction network $\epsilon_\text{rgb}$, which is computationally expensive, we optimize the decoder $\mathcal{D}$ with respect to the input image $I_\text{o}$. This decoder fine-tuning strategy helps preserve the visible pixel information with minimal computational overhead. The fine-tuning objective is defined as:
\begin{multline}
\label{eq16}
    \theta^* = \operatorname*{argmin}_{\theta} ( 
    \| \mathcal{D}_{\theta}(z_\text{do}) \odot (1 - M_\text{m}) - I_\text{do} \odot (1 - M_\text{m}) \| \\
    + \lambda \| \mathcal{D}_{\theta}(z_\text{do}) \odot M_\text{m} - I_\text{o} \odot M_\text{m} \| ),
\end{multline}
where $\odot$ denotes element-wise multiplication, and $\theta^*$ denotes the optimized decoder parameters obtained via fine-tuning. $\lambda$ is a hyperparameter that weights visible region preservation during optimization. With the decoder fine-tuned via Eq.~\eqref{eq16}, the final de-occluded image $I_\text{do}^*$ is obtained by decoding $z_\text{do}$ as:
\begin{equation}
\label{eq17}
I_\text{do}^* = \mathcal{D}_{\theta^*} (z_\text{do}),
\end{equation}
where $I_\text{do}^*$ exhibits improved preservation of visible regions compared to $I_\text{do} = \mathcal{D}_\theta(z_\text{do})$ obtained from the original decoder without fine-tuning.

\begin{table}[t]
    \centering
    {\scriptsize
    \renewcommand{\arraystretch}{1.2}
    \setlength{\tabcolsep}{3pt}
    \begin{tabular}{l|ccc|ccc}
        \toprule
        \raisebox{-2ex}{Method} & \multicolumn{3}{c|}{AHP real dataset} & \multicolumn{3}{c}{AHP syn dataset} \\ \cline{2-4} \cline{5-7}
        & mIoU $\uparrow$ & mIoU-inv $\uparrow$ & L1 $\downarrow$
        & mIoU $\uparrow$ & mIoU-inv $\uparrow$ & L1 $\downarrow$ \\ \midrule
SSSD~\cite{zhan2020self}\dag  & 81.3 & 31.2 & 0.222 & 83.1 & 29.1 & 0.196 \\ 
Human DO~\cite{zhou2021human}\dag & 86.1 & 40.3 & 0.164 & 84.6 & 43.7 & 0.150 \\
amodal~\cite{xu2024amodal} & 85.1 & \underline{57.1} & 0.148 & 77.4 & 40.2 & 0.333 \\
pix2gestalt~\cite{ozguroglu2024pix2gestalt}  & 90.0 & 54.8 & 0.101 & 91.9 & \underline{65.6} & \underline{0.108} \\
SDAmodal~\cite{zhan2024amodal} & \underline{91.6} & 56.1 & \underline{0.087} & \underline{92.3} & 65.5 & 0.109 \\
Ours  & \textbf{93.5} & \textbf{65.1} & \textbf{0.070} & \textbf{92.6} & \textbf{67.1} & \textbf{0.105} \\
        \bottomrule
    \end{tabular}
    }
    \caption{Quantitative comparison results for mask completion. The best performance for each metric is highlighted in \textbf{bold}, while the second-best is \underline{underlined}. \dag~denotes values reported in~\cite{zhou2021human}.}
    \vspace{-8mm}
    \label{table:1}
\end{table}

\section{Experimental Results}

\subsection{Datasets}
\label{subsec:datasets}

For training, we constructed a dataset comprising 55,347 occluded human examples, each containing a modal mask, an amodal mask, an occluded RGB image, and its corresponding non-occluded RGB image. For evaluation, we used 891 synthetic and 56 real test images, both obtained from the AHP dataset~\cite{zhou2021human}. The training data were generated by applying synthetic occlusions to non-occluded human images from the AHP dataset, using randomly selected objects from the COCOA dataset~\cite{zhu2017semantic,follmann2019learning}. To increase diversity, we applied data augmentation techniques such as color jittering, random shifts, and horizontal flipping. The occlusion ratio was sampled from a Gaussian distribution to approximate realistic occlusion patterns. Further details on the training dataset are described Section A.5 of the supplementary material.

\subsection{Implementation Details}

Our proposed network requires no additional training beyond the mask completion U-Net, $\mathcal{U}_\text{mask}$, as all other components rely on pre-trained models. To extract the diffusion-based human body prior, we adopt the denoising U-Net and ControlNet weights from~\cite{zhu2024dpmesh}, which fine-tune Stable Diffusion~\cite{rombach2022high} for SMPL~\cite{2015_SMPL} parameter estimation. The mask completion U-Net $\mathcal{U}_\text{mask}$ is trained using the stochastic gradient descent optimizer with a learning rate of $1\times10^{-3}$ and a batch size of 16 for 14,000 iterations on two RTX 4090 GPUs, requiring approximately two and a half hours to complete.

For 2D pose estimation, we used OpenPose~\cite{8765346} to extract 25 joint coordinates, denoted as $J_\text{2D} \in \mathbb{R}^{k \times 2}$, where $k = 25$. To increase spatial resolution, each joint pair was uniformly divided into 10 subdivisions, resulting in 9 sub-joints between each pair of joints. This process was applied to 14 selected joint pairs, yielding a total of $14 \times 9 = 126$ sub-joints, denoted as $J_\text{sub} \in \mathbb{R}^{s \times 2}$, where $s = 126$. We excluded six foot-related joints and retained the remaining 19 body joints, denoted as $J_\text{body} \in \mathbb{R}^{19 \times 2}$. The final dense joint set $J_\text{dense} \in \mathbb{R}^{145 \times 2}$ was formed by concatenating $J_\text{body}$ and $J_\text{sub}$. To generate the joint heatmaps, we set the standard deviation of the Gaussian kernel to $\sigma_1 = 2.5$ for the 2D joint heatmap $H_\text{2D}$, and $\sigma_2 = 8$ for the occluded joint heatmap $H_\text{o}$.

For human-specific text feature extraction, we used BLIP~\cite{li2022blip} as VQA model, and for RGB completion, we adopted the denoising U-Net from~\cite{ozguroglu2024pix2gestalt}, which fine-tuned Stable Diffusion for the de-occlusion task. To condition the RGB completion on the structural information provided by the amodal mask $M_\text{a}$, we utilized the \emph{control\_v11p\_sd15\_segmentation} checkpoint of ControlNet to extract the control feature $f_\text{c2}$. Decoder fine-tuning was performed for 50 steps, requiring approximately 14 seconds per image. The hyperparameter $\lambda$ was set to 100, following~\cite{avrahami2023blended}.

\subsection{Evaluation Metrics}

To evaluate the performance of the proposed method, we employ task-specific metrics. For mask completion, we report overall accuracy using Mean Intersection over Union (mIoU) and L1 loss. In addition, inverse mIoU (mIoU-inv) is used to specifically assess performance in occluded regions. In this setting, mIoU is computed over the entire mask area, while mIoU-inv is calculated only within the occluded regions by excluding the modal mask area from the amodal GT.

For RGB completion, we assess the quality of reconstructed images using Peak Signal-to-Noise Ratio (PSNR), Learned Perceptual Image Patch Similarity (LPIPS), L1 loss, and Fréchet Inception Distance (FID). Among these, PSNR and L1 loss focus on pixel-level reconstruction accuracy, whereas LPIPS and FID evaluate perceptual quality and visual realism.

\begin{table}[t]
    \centering
    {\scriptsize
    \renewcommand{\arraystretch}{1.2}
    \setlength{\tabcolsep}{3pt}
    \begin{tabular}{l|cccc|cccc}
        \toprule
        \raisebox{-2ex}{Method} & \multicolumn{4}{c|}{AHP real dataset} & \multicolumn{4}{c}{AHP syn dataset} \\
        \cline{2-5} \cline{6-9}
        & L1 $\downarrow$ & FID $\downarrow$ & LPIPS* $\downarrow$ & PSNR $\uparrow$
        & L1 $\downarrow$ & FID $\downarrow$ & LPIPS* $\downarrow$ & PSNR $\uparrow$ \\ \midrule
SSSD~\cite{zhan2020self}\dag & 0.0911 & 28.30 & - & - & 0.0936 & 18.50 & - & - \\
Human DO~\cite{zhou2021human}\dag & 0.0617 & 19.49 & - & - & 0.0519 & 13.85 & - & - \\
amodal~\cite{xu2024amodal} & 0.0197 & 16.83 & 56.24 & 21.13 & 0.0232 & 13.59 & 72.11 & 19.74 \\
pix2gestalt~\cite{ozguroglu2024pix2gestalt} & \underline{0.0100} & \underline{10.34} & \underline{47.73} & \underline{25.88} & \underline{0.0124} & \underline{10.03} & \underline{51.76} & \underline{25.18} \\
Ours & \textbf{0.0062} & \textbf{8.26} & \textbf{37.63} & \textbf{29.13} & \textbf{0.0072} & \textbf{5.46} & \textbf{27.86} & \textbf{28.36} \\
        \bottomrule
    \end{tabular}
    }
    \caption{Quantitative comparison results for RGB completion. LPIPS* denotes LPIPS$\times$1000 for readability.}
    \vspace{-8mm}
    \label{table:2}
\end{table}

\begin{figure}
    \centering
    \includegraphics[width=0.9\linewidth]{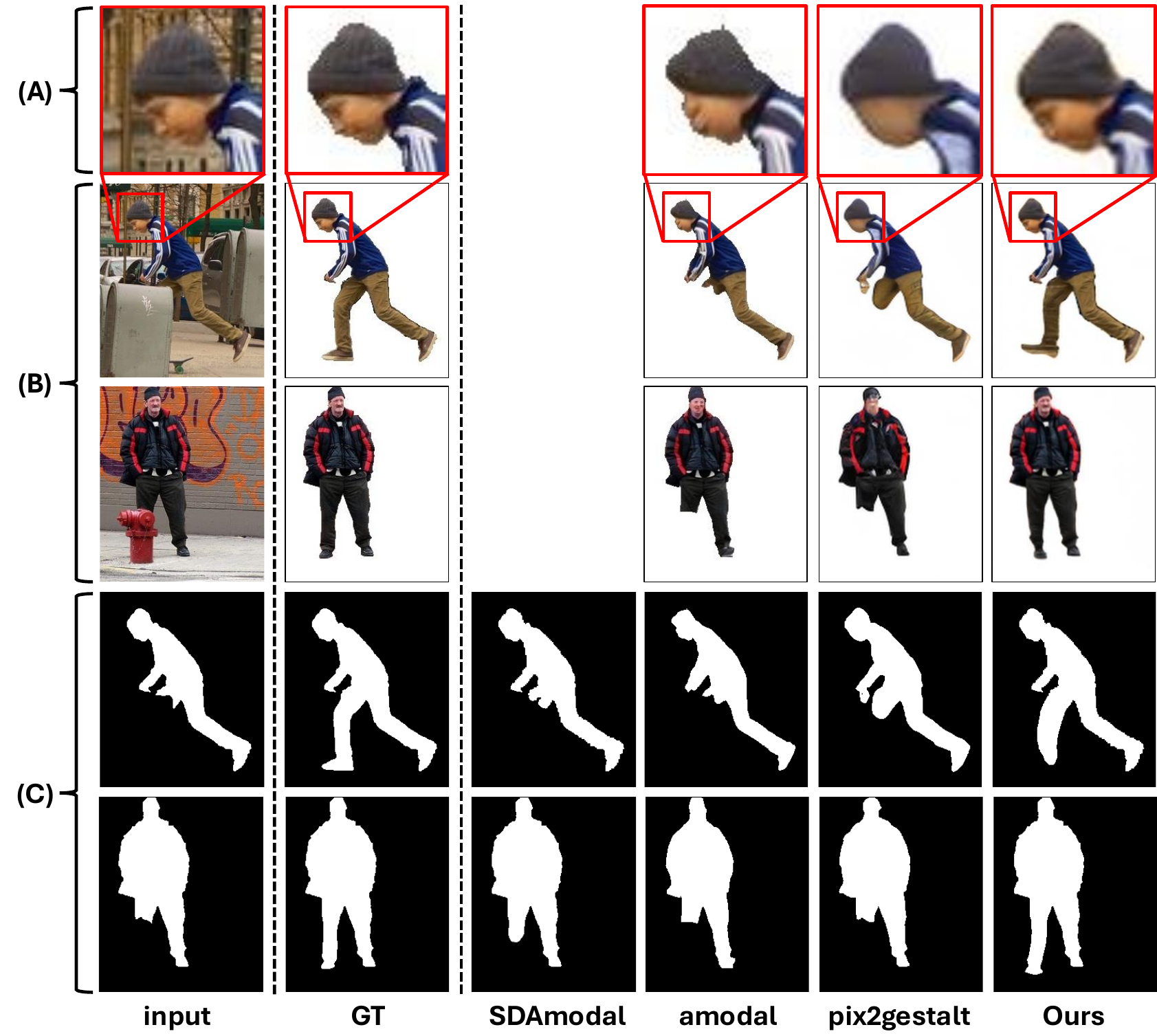}
    \vspace{-4mm}
    \caption{Qualitative results on the AHP real dataset. (A): partial RGB results, (B): RGB completion, (C): mask completion.}
    \label{fig:4}
    \vspace{-5mm}
\end{figure}

\subsection{Comparison with Existing Methods}

We evaluate the de-occlusion performance of our model in two aspects: mask completion and RGB completion. For both tasks, we compare our method with existing de-occlusion models~\cite{ozguroglu2024pix2gestalt,xu2024amodal}. Note that the method in~\cite{zhan2020self} requires an additional input mask indicating the occluding object, and the approaches in~\cite{zhou2021human,liu2024object} cannot be directly compared due to the lack of publicly available implementations. To enable a fair comparison, we include the results reported in~\cite{zhou2021human} on the same AHP dataset. In addition, we compare our mask completion network with the current state-of-the-art method~\cite{zhan2024amodal} for amodal completion.

Table~\ref{table:1} presents the quantitative performance of our method on the mask completion task, demonstrating that it outperforms all existing models across all metrics on both the AHP real and synthetic datasets. The improvement is particularly notable in the occluded regions of real-world images, which are generally more challenging. Table~\ref{table:2} shows that our method also achieves superior performance on RGB completion, significantly surpassing previous approaches. Note that~\cite{zhan2020self} and~\cite{zhou2021human} report only L1 and FID scores in~\cite{zhou2021human}, so the remaining metrics could not be included in our comparison.

Qualitative results in Fig.~\ref{fig:4} further demonstrate the effectiveness of our method in handling human occlusions. For example, when a human leg is occluded, humans can intuitively infer its presence based on physical plausibility (e.g., standing posture). However, existing amodal completion and de-occlusion methods often oversmooth the boundaries and fail to accurately reconstruct the missing body part. In contrast, our method utilizes human-specific priors to more effectively complete the amodal mask, including occluded legs, and subsequently recover the corresponding RGB appearance. Since SDAmodal performs only mask completion, it does not generate de-occluded RGB images; thus, its RGB results are omitted from Fig.~\ref{fig:4}. Moreover, Stable Diffusion-based methods, such as the amodal~\cite{xu2024amodal} and pix2gestalt~\cite{ozguroglu2024pix2gestalt}, frequently introduce distortions or degrade visible regions, especially in the face area, as shown in (A) of Fig.~\ref{fig:4}. In contrast, our method better preserves visible details, generating outputs that are visually consistent with the input image.

\subsection{Ablation Experiments}
\label{subsec:ab}

In this section, we present ablation experiments that validate the architectural choices of the proposed method.

\noindent\textbf{Mask completion network.} To evaluate the effectiveness of our proposed human priors, we conducted an ablation study on the AHP real dataset by incrementally adding each component, as shown in Table~\ref{table:3}. Starting from a baseline configuration (\#1) without any human prior, we sequentially introduced the diffusion-based human body prior $\mathcal{F}, \mathcal{T}$ in (\#2), the occluded joint heatmap $H_{o}$ in (\#3), and the subdivided joints $J_\text{sub}$ in (\#4).

The diffusion-based human body prior improved both mIoU and mIoU-inv, but increased the L1 loss. This is likely due to pixel-level misalignment or over-smoothing, despite better structural estimation. Incorporating the occluded joint heatmap addressed this issue by providing direct localization of occluded regions and led to performance gains across all metrics. Finally, adding the subdivided joints further enhanced results by increasing the spatial density of pose guidance.

\begin{figure}
    \centering
    \includegraphics[width=0.9\linewidth]{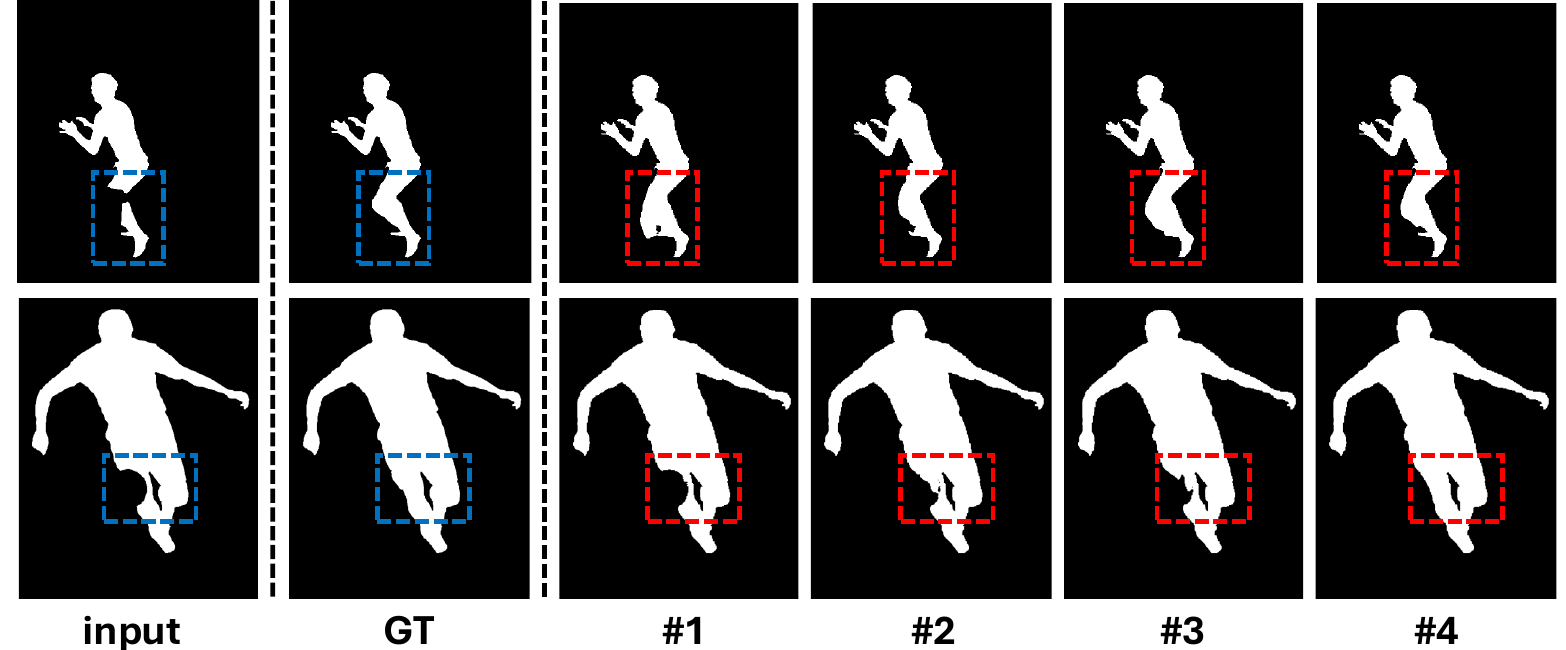}
    \vspace{-4mm}
    \caption{Qualitative ablation results on the mask completion network. \#1–\#4 correspond to the ID settings in Table~\ref{table:3}.}
    \label{fig:5}
    \vspace{-5mm}
\end{figure}

Qualitative results corresponding to this ablation study are presented in Fig.~\ref{fig:5}. The baseline configuration (\#1) often fails to reconstruct occluded regions or tends to over-reconstruct missing parts. The introduction of the diffusion-based human body prior and occluded joint heatmap (\#2, \#3) improves structural plausibility, although some limitations remain. The final configuration (\#4), which integrates all proposed components, successfully reconstructs occluded regions and produces amodal masks that are visually more consistent with the GT.

\begin{table}
    \centering
    {\scriptsize
    \renewcommand{\arraystretch}{1.2}
    \begin{tabular}{c|ccc|ccc}
        \toprule
        ID & $\mathcal{F},\mathcal{T}$ & $H_\text{o}$ & $J_\text{sub}$ & mIoU $\uparrow$ & mIoU-inv $\uparrow$ & L1 $\downarrow$ \\
        \midrule
        \#1 &   &       &       & 92.3 & 58.8 & 0.077 \\
        \#2 & \checkmark &       &       & 92.4 & 60.3 & 0.082 \\
        \#3 & \checkmark & \checkmark &       & 92.8 & 62.2 & 0.076 \\
        \#4 & \checkmark & \checkmark & \checkmark & \textbf{93.5} & \textbf{65.1} & \textbf{0.070} \\
        \bottomrule
    \end{tabular}
    }
    \caption{Quantitative ablation results on the mask completion network. $\mathcal{F}, \mathcal{T}$ denote the diffusion-based human body prior, $H_\text{o}$ represents the occluded joint heatmap, and $J_\text{sub}$ indicates the subdivided joints.}
    \vspace{-6mm}
    \label{table:3}
\end{table}

\noindent\textbf{RGB completion network.} We conducted an ablation study on the RGB completion network using the AHP real dataset to evaluate the contribution of each proposed component, as summarized in Table~\ref{table:4}. Using the control feature $f_\text{c2}$ extracted from the amodal mask (\#6) improved all evaluation metrics except FID, compared to the no-guidance baseline (\#5), even without incorporating additional semantic information. This indicates that structural localization alone can effectively guide the reconstruction of occluded regions. However, the drop in FID suggests limitations in generating perceptually realistic content. To address this, we incorporated the human-specific text feature $f_\text{hs}$ (\#7), which provides semantic context related to human appearance. This addition not only mitigated the FID degradation but also yielded consistent improvements across all metrics. Finally, fine-tuning the decoder parameters $\theta^*$ (\#8) further enhanced reconstruction fidelity by reducing artifacts and better preserving details in the visible regions. These results highlight the complementary roles of spatial and semantic conditioning in achieving high-quality RGB completion.

\begin{figure}
    \centering
    \includegraphics[width=0.9\linewidth]{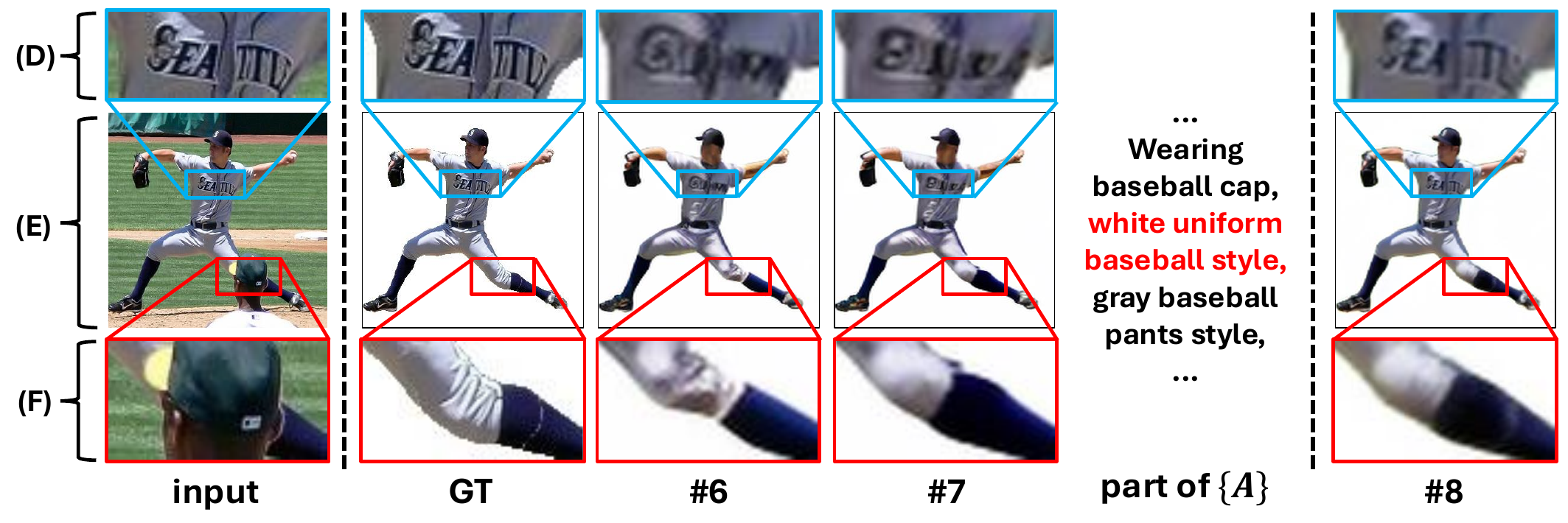}
    \vspace{-4mm}
    \caption{Qualitative ablation results on the RGB completion network. (D) and (F): partial RGB results, (E): full RGB results. \#6–\#8 correspond to the ID settings in Table~\ref{table:4}.}
    \label{fig:6}
    \vspace{-4mm}
\end{figure}

As shown in (F) of Fig.~\ref{fig:6}, configuration (\#6), which does not use the human-specific text feature $f_\text{hs}$, often fails to handle occlusions effectively, resulting in artifacts in the occluded areas. By incorporating $f_\text{hs}$, extracted from the human-specific description $\{A\}$, configuration (\#7) can mitigate such artifacts and better reconstruct occluded content. Specifically, the red sentence in $\{A\}$ provides semantic cues that guide the generation process, enabling more accurate synthesis of the missing parts based on the visible context. Finally, decoder fine-tuning in configuration (\#8) further improves reconstruction quality. As highlighted in (D) of Fig.~\ref{fig:6}, both \#6 and \#7 fail to recover the word "Seattle", whereas \#8 successfully reconstructs it in a manner closely resembling the GT. Additional ablation studies on RGB completion are provided in the supplementary material.

\begin{table}
    \centering
    {\scriptsize
    \renewcommand{\arraystretch}{1.2}
    \begin{tabular}{c|ccc|cccc}
        \toprule
        ID & $f_\text{c2}$ & $f_\text{hs}$ & $\theta^*$ & L1 $\downarrow$ & FID $\downarrow$ & LPIPS* $\downarrow$ & PSNR $\uparrow$ \\
        \midrule
        \#5 &   &   &   & 0.0100 & 10.34 & 47.73 & 25.88 \\
        \#6 & \checkmark &   &   & 0.0099 & 10.85 & 47.64 & 25.92 \\
        \#7 & \checkmark & \checkmark &   & 0.0096 & 9.86 & 47.27 & 26.27 \\
        \#8 & \checkmark & \checkmark & \checkmark & \textbf{0.0062} & \textbf{8.26} & \textbf{37.63} & \textbf{29.13} \\
        \bottomrule
    \end{tabular}
    }
    \caption{Quantitative ablation results on the RGB completion network. $f_\text{c2}$ denotes the control feature derived from the amodal mask $M_\text{a}$, $f_\text{hs}$ represents the human-specific text feature, and $\theta^*$ indicates the optimized decoder parameters.}
    \vspace{-6mm}
    \label{table:4}
\end{table}

\noindent\textbf{Two-stage framework design.}
Through quantitative evaluation, we show that decoupling mask completion and RGB completion within a two-stage baseline leads to improved performance. For comparison, we implement a one-stage framework that jointly predicts the amodal mask $M_\text{a}$ and the de-occluded RGB image $I_\text{do}$, formulated as:
\begin{equation}
\label{eq18}
    M_\text{a}, I_\text{do} = \mathcal{U}_\text{one}(M_\text{m} \oplus H_\text{o} \oplus I_\text{o}, \mathcal{F}, \mathcal{T}),
\end{equation}
where $\mathcal{U}_\text{one}$ is derived by extending the input and output structure of the mask completion U-Net $\mathcal{U}_\text{mask}$. Here, $I_\text{o}$, $M_\text{m}$, and $H_\text{o}$ indicate the input image, modal mask, and occluded joint heatmap.

As shown in Table~\ref{table:5}, the two-stage framework consistently outperforms the one-stage baseline in both mask and RGB completion on the AHP synthetic dataset. This result highlights the effectiveness of using the amodal mask $M_\text{a}$ as a conditioning input and enabling each network to specialize in its respective sub-task. Since the mask completion network remains fixed across all two-stage variants, their performance on the mask completion task remains unchanged.

\noindent\textbf{Stage-specific architectural design.}
To construct a two-stage framework, we first consider a baseline referred to as two-stage ($\mathcal{U}_\text{rgb}$)\ddag{}, in which the RGB image is reconstructed using a network based on the same architecture as the mask completion U-Net $\mathcal{U}_\text{mask}$, with appropriately modified inputs and outputs.

In this variant, we replace the diffusion-based human body prior $\mathcal{F}$ and $\mathcal{T}$ from $\epsilon_\text{mask}$ with the RGB-based diffusion prior extracted from $\epsilon_\text{rgb}$, denoted as $\mathcal{F}^\prime$ and $\mathcal{T}^\prime$, formulated as:
\begin{equation}
\label{eq19}
    \mathcal{F}^\prime, \mathcal{T}^\prime = \epsilon_\text{rgb}(z_0^\prime, t, \mathcal{C}_2^\prime),
\end{equation}
where $\mathcal{C}_2^\prime = \{f_\text{lw}, f_\text{c2}\}$ is a conditioning input consisting of a lightweight text feature $f_\text{lw} = \mathcal{E}_\text{clip}(I_\text{o})$ and a control feature $f_\text{c2}$ derived from the amodal mask $M_\text{a}$. The lightweight text feature $f_\text{lw}$ replaces the original human-specific text feature $f_\text{hs}$ to simplify the experimental setup. The diffusion timestep $t$ is set to 0 to ensure that no noise is added to the latent feature map.

Using the diffusion priors $\mathcal{F}^\prime$ and $\mathcal{T}^\prime$, the RGB completion U-Net $\mathcal{U}_\text{rgb}$ reconstructs the de-occluded image $I_\text{do}$ as follows:
\begin{equation}
\label{eq20}
    I_\text{do} = \mathcal{U}_\text{rgb}(I_\text{o} \oplus M_\text{m} \oplus M_\text{i}, \mathcal{F}^\prime, \mathcal{T}^\prime),
\end{equation}
where $M_\text{i} = M_\text{a} - M_\text{m}$ denotes the occluded region mask. 

As shown in Table~\ref{table:5}, although the two-stage ($\mathcal{U}_\text{rgb}$)\ddag{} variant outperforms the one-stage baseline, its improvements in perceptual quality metrics such as FID and LPIPS are relatively limited. This suggests that simple feedforward extraction of diffusion priors may be insufficient for high-fidelity RGB reconstruction.

To address this limitation, we propose performing RGB completion through an iterative denoising process, rather than relying solely on extracted priors. The resulting two-stage (Ours)\ddag{} model consistently outperforms the two-stage ($\mathcal{U}_\text{rgb}$)\ddag{} variant, particularly in perceptual quality metrics. Furthermore, the performance gap between two-stage (Ours)\ddag{} and two-stage (Ours) highlights the benefit of incorporating the human-specific text feature $f_\text{hs}$.

\begin{table}
    \centering
    {\scriptsize
    \renewcommand{\arraystretch}{1.2}
    \setlength{\tabcolsep}{3pt}
    \begin{tabular}{l|ccc|cccc}
        \toprule
        \raisebox{-2ex}{Method} & \multicolumn{3}{c|}{Mask Completion} & \multicolumn{4}{c}{RGB Completion} \\
        \cline{2-4} \cline{5-8}
        & mIoU $\uparrow$ & mIoU-inv $\uparrow$ & L1 $\downarrow$ & L1 $\downarrow$ & FID $\downarrow$ & LPIPS* $\downarrow$ & PSNR $\uparrow$ \\
        \midrule
        one-stage & 90.3 & 58.3 & 0.141 & 0.0084 & 8.23 & 35.74 & 28.16 \\
        two-stage ($\mathcal{U}_\text{rgb}$)\ddag & \textbf{92.6} & \textbf{67.1} & \textbf{0.105} & 0.0080 & 7.92 & 32.51 & 28.35 \\
        two-stage (Ours)\ddag & \textquotedbl & \textquotedbl & \textquotedbl & 0.0073 & 5.92 & 27.89 & 28.30 \\
        two-stage (Ours) & \textquotedbl & \textquotedbl & \textquotedbl & \textbf{0.0072} & \textbf{5.46} & \textbf{27.86} & \textbf{28.36} \\
        \bottomrule
    \end{tabular}
    }
    \caption{Quantitative comparison between one-stage and two-stage frameworks, including stage-specific design choices. \ddag{} denotes the use of $f_\text{lw}$ instead of $f_\text{hs}$.}
    \vspace{-6mm}
    \label{table:5}
\end{table}

\section{Conclusion}

We presented a two-stage framework for human de-occlusion that separately addresses mask completion and RGB completion. Our method leverages structural priors through diffusion-based human body representations and occluded joint heatmaps, and incorporates appearance-level priors using human-specific text features extracted from a VQA model. The amodal mask predicted in the first stage serves as an explicit guide for RGB reconstruction, while decoder fine-tuning enhances the preservation of visible details. Extensive experiments demonstrate that our approach consistently outperforms existing methods in both mask and RGB completion. Furthermore, the de-occluded images generated by our framework can be effectively utilized as enhanced inputs for downstream tasks such as 2D pose estimation, 3D clothed human reconstruction, and 3D human mesh recovery, leading to improved task performance.

\begin{acks}
This work was supported by Institute of Information \& Communications Technology Planning \& Evaluation (IITP) grant funded by the Korea government (MSIT) (No. RS-2023-00219700, Development of FACS-compatible Facial Expression Style Transfer Technology for Digital Human).
\end{acks}

\appendix
\setcounter{section}{0}
\setcounter{equation}{0}
\setcounter{figure}{0}
\setcounter{table}{0}

\renewcommand{\theequation}{\alph{equation}}
\renewcommand{\thesection}{\Alph{section}}
\renewcommand{\thefigure}{\Alph{figure}}
\renewcommand{\thetable}{\Alph{table}}

\makeatletter
\twocolumn[%
  \begin{center}
    {\@titlefont Supplementary Materials: Stable Diffusion-Based Approach for Human De-Occlusion}
    \vspace*{8mm}
  \end{center}
]
\makeatother

\section{Detailed Implementations}

This section presents the implementation details of our method, including the generation of occluded joint heatmaps, the set of predefined VQA questions, and the loss functions used to train the ablation models. Additionally, we describe the design choice of using ControlNet for injecting human body priors into the diffusion models, as well as details on the constructed training dataset.

\subsection{Occluded Joint Heatmap}
\label{suppl_subsec:a1}

\begin{figure}
    \centering
    \includegraphics[width=\linewidth]{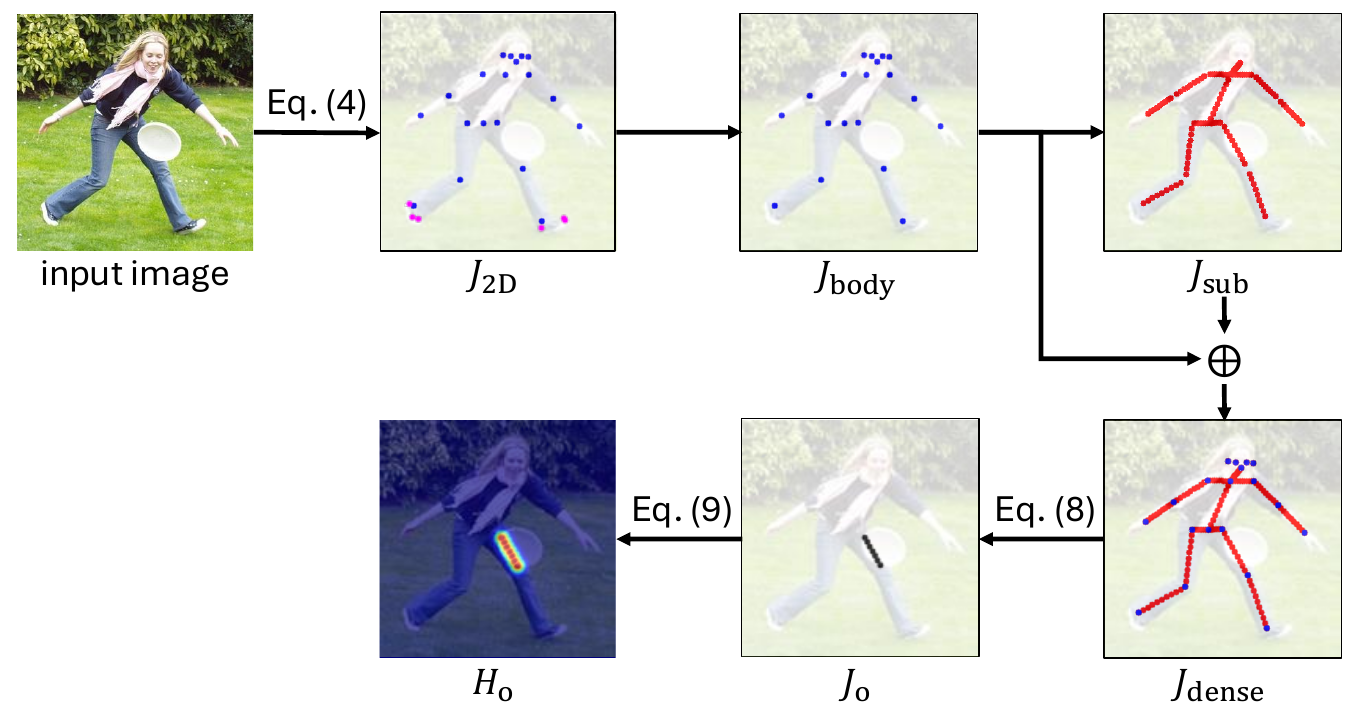}
    \vspace{-6mm}
    \caption{Occluded joint heatmap generation process.}
    \label{suppl_fig:a}
\end{figure}

We provide a detailed explanation of how the occluded joint heatmap, denoted as $H_\text{o}$, is generated from an input image, as illustrated in Fig.~\ref{suppl_fig:a}.

Given an input image, we first extract 25 2D joints $J_{\text{2D}} \in \mathbb{R}^{25 \times 2}$ using OpenPose~\cite{8765346}. We remove six foot joints---highlighted in magenta in Fig.~\ref{suppl_fig:a}---to obtain the body joint set $J_{\text{body}} \in \mathbb{R}^{19 \times 2}$, visualized in blue. Based on $J_{\text{body}}$, we interpolate 14 predefined joint pairs to construct the subdivided joints $J_{\text{sub}} \in \mathbb{R}^{126 \times 2}$, shown in red. Each joint pair is divided into 10 equal segments, resulting in 9 interpolated joints per pair. These joint pairs follow standard definitions commonly used in 2D human pose estimation, as listed in Table~\ref{suppl_tab:a}.

We then concatenate $J_{\text{body}}$ and $J_{\text{sub}}$ to form a dense joint representation $J_{\text{dense}} \in \mathbb{R}^{145 \times 2}$, where $J_{\text{dense}} = J_{\text{body}} \oplus J_{\text{sub}}$. To identify the joints located in occluded regions, we follow the procedure described in Eq.~(8) of the main paper. The resulting occluded joint set, $J_\text{o}$, is visualized in black and contains only joints within occluded areas. Finally, $J_\text{o}$ is converted into an occluded joint heatmap $H_\text{o}$ by applying a 2D Gaussian kernel, as defined in Eq.~(9).

\begin{table}
    \centering
    \renewcommand{\arraystretch}{1.2}
    \begin{tabular}{c|c|c}
        \toprule
        Index & Joint Pair & Joint Names \\
        \midrule
        1 & (0, 1) & Nose $\rightarrow$ Neck \\
        2 & (1, 2) & Neck $\rightarrow$ RShoulder \\
        3 & (2, 3) & RShoulder $\rightarrow$ RElbow \\
        4 & (3, 4) & RElbow $\rightarrow$ RWrist \\
        5 & (1, 5) & Neck $\rightarrow$ LShoulder \\
        6 & (5, 6) & LShoulder $\rightarrow$ LElbow \\
        7 & (6, 7) & LElbow $\rightarrow$ LWrist \\
        8 & (1, 8) & Neck $\rightarrow$ Pelvis \\
        9 & (8, 9) & Pelvis $\rightarrow$ RHip \\
        10 & (9, 10) & RHip $\rightarrow$ RKnee \\
        11 & (10, 11) & RKnee $\rightarrow$ RAnkle \\
        12 & (8, 12) & Pelvis $\rightarrow$ LHip \\
        13 & (12, 13) & LHip $\rightarrow$ LKnee \\
        14 & (13, 14) & LKnee $\rightarrow$ LAnkle \\
        \bottomrule
    \end{tabular}
    \vspace{1mm}
    \caption{Predefined joint pairs used for interpolation to construct the subdivided joints $J_\text{sub}$. Each pair connects two joints according to the OpenPose joint index.}
    \label{suppl_tab:a}
    \vspace{-6mm}
\end{table}

\subsection{VQA Questions}

To obtain the human-specific text feature $f_\text{hs}$, we apply a set of fine-grained VQA questions $\{Q\}$ to the input image. The model generates a set of descriptive answers $\{A\}$ that capture garment and appearance attributes. These descriptions are then encoded using a CLIP text encoder $\mathcal{E}_\text{clip}$ to produce the feature $f_\text{hs}$. This section details the composition of the predefined question set $\{Q\}$.

The set $\{Q\}$ is largely based on the TeCH~\cite{huang2024tech} framework, with minor modifications. Specifically, questions related to facial appearance (e.g., “Describe the facial appearance”) were excluded from $Q_{\text{appearance}}$ due to poor response quality from the BLIP model (e.g., generic outputs like “white”). In contrast, garment-related questions in $Q_{\text{garments}}$ were retained without modification. The complete set of questions is listed in Table~\ref{suppl_tab:b}.

Following TeCH, we define a garment category set $\mathcal{G}$ containing ten categories: hat, sunglasses, upper-clothes, skirt, pants, dress, belt, shoes, bag, and scarf. For each garment $g \in \mathcal{G}$, the questions from $Q_\text{garments}$ in Table~\ref{suppl_tab:b} are applied. If the initial question ("Is this person wearing $g$?") is answered affirmatively ($a_2$ is yes), follow-up questions are issued. Similarly, for facial hair, the follow-up question is asked only when the previous answer $a_1$ indicates the presence of facial hair.

To address redundancy in VQA outputs, we apply post-processing to remove repeated words and concatenate the cleaned responses to form the final sentence set $\{A\}$, which serves as input to the CLIP encoder.

\begin{table}
    \centering
    \renewcommand{\arraystretch}{1.2}
    \begin{tabular}{ll}
    \toprule
        Groups & Questions $\{Q\}$ \\
        \midrule
        \multirow{7}{*}{$Q_\text{appearance}$} & Is this person a man or a woman? \\
        & What is the hair color of this person? \\
        & What is the hairstyle of this person? \\
        & Does this person have facial hair? $\rightarrow a_1$ \\
        & If $a_1$ is yes: How is the facial hair of this person? \\
        & Describe the pose of this person. \\
        \midrule
        \multirow{4}{*}{$Q_\text{garments}$} 
        & Is this person wearing $g$? $\rightarrow a_2$ \\
        & If $a_2$ is yes: What $g$ is the person wearing? $\rightarrow k$ \\
        & If $a_2$ is yes: What is the color of the $k$ + $g$? \\
        & If $a_2$ is yes: What is the style of the $k$ + $g$? \\
        \bottomrule
    \end{tabular}
    \vspace{1mm}
    \caption{Predefined VQA question set $\{Q\}$. Here, $g$ denotes a garment category, and $k$ refers to the garment kind inferred by the VQA model from the second question in $Q_\text{garments}$. Follow-up questions are included only if the preceding answer---denoted as $a_1$ or $a_2$---is affirmative.}
    \label{suppl_tab:b}
    \vspace{-8mm}
\end{table}

\subsection{Ablation Models}

This section outlines the loss functions used to train the ablation models designed to validate our architectural choices.

The one-stage model integrates both mask and RGB completion within a single U-Net architecture $\mathcal{U}_{\text{one}}$, jointly predicting the amodal mask $M_\text{a}$ and the de-occluded image $I_\text{do}$. For mask prediction, we use the binary cross-entropy (BCE) loss $\mathcal{L}_{\text{BCE}}$, consistent with the formulation in Eq.~(11) of the main paper. For RGB completion, we adopt the inpainting loss from SSSD~\cite{zhan2020self}, which combines several loss components computed over both visible and occluded regions:
\begin{multline}
\label{eq_inpaint}
    \mathcal{L}_{\text{inpaint}} = 
    \lambda_{\text{invis}} \mathcal{L}_{\text{invis}} + \lambda_{\text{vis}} \mathcal{L}_{\text{vis}} \\
    + \lambda_{\text{prec}} \mathcal{L}_{\text{prec}}
    + \lambda_{\text{style}} \mathcal{L}_{\text{style}} + \lambda_{\text{tv}} \mathcal{L}_{\text{tv}},
\end{multline}
\noindent where $\mathcal{L}_{\text{invis}}$ and $\mathcal{L}_{\text{vis}}$ are L1 losses on invisible and visible regions, respectively. $\mathcal{L}_{\text{prec}}$ is a perceptual loss based on a pretrained VGG network, $\mathcal{L}_{\text{style}}$ is a style loss using Gram matrices, and $\mathcal{L}_{\text{tv}}$ is the total variation loss promoting smoothness within the foreground. Following the hyperparameter configuration from SSSD, we set $\lambda_{\text{invis}}=6$, $\lambda_{\text{vis}}=1$, $\lambda_{\text{prec}}=0.1$, $\lambda_{\text{style}}=250$, and $\lambda_{\text{tv}}=0.1$.

The final loss for training $\mathcal{U}_{\text{one}}$ is:
\begin{equation}
    \mathcal{L}_{\text{total}} = \lambda_{\text{BCE}} \mathcal{L}_{\text{BCE}} + \mathcal{L}_{\text{inpaint}},
\end{equation}
where $\lambda_{\text{BCE}}$ is set to 10 to balance the scale between $\mathcal{L}_{\text{BCE}}$ and $\mathcal{L}_{\text{inpaint}}$.

The two-stage variant $\mathcal{U}_{\text{rgb}}$ uses the same inpainting loss $\mathcal{L}_{\text{inpaint}}$ as in Eq.~(\ref{eq_inpaint}) for RGB completion training. Unlike the one-stage model, $\mathcal{U}_{\text{rgb}}$ is trained only for RGB completion after the mask completion stage is completed. Therefore, it does not include the BCE loss component.

\subsection{ControlNet Design Choice}
\label{suppl_subsec:a4}

In our framework, we employ ControlNet~\cite{zhang2023adding} to incorporate human body priors (HBP) into the diffusion model (DM) used for both mask and RGB completion. This design choice allows structural guidance to be injected without the need to modify or retrain the base DM. HBP can be integrated into DM in three main ways: (1) training a new DM architecture that takes HBP as additional input; (2) fine-tuning a pre-trained DM to integrate HBP; and (3) injecting HBP using ControlNet. 

The first two approaches require training or fine-tuning large DM, which is time-consuming and computationally expensive. In contrast, ControlNet enables the injection of HBP without retraining the DM. By using ControlNet, our framework updates only a small subset of parameters: 18.02M out of 1444.33M (1.25\%) for the mask completion and 49.49M out of 2370.49M (2.09\%) for the RGB completion.

\subsection{Train Dataset}
\label{suppl_subsec:a5}

As described in Section 4.1 of the main paper, we constructed the training dataset by combining the AHP dataset~\cite{zhou2021human} and the COCOA dataset~\cite{zhu2017semantic, follmann2019learning}. From the training split of the AHP dataset, we obtained unoccluded human RGB images along with their corresponding whole-body segmentation masks. From the COCOA dataset, we sampled in-the-wild objects with associated RGB regions and segmentation masks.

To simulate occlusions, we synthetically overlaid COCOA objects onto the unoccluded AHP human images. The degree of occlusion was controlled by a predefined occlusion ratio. Objects were randomly placed over the human region and iteratively repositioned until the predefined occlusion ratio was satisfied. If the desired occlusion ratio was still unmet after a fixed number of iterations, a new object was sampled from the COCOA dataset, and the placement procedure was restarted.

Since the occlusions are synthetic, any mask errors stem from the original annotations of AHP or COCOA. While the COCOA dataset provides manually annotated masks, the AHP dataset contains masks that were either manually labeled (e.g., COCO, VOC, SBD, LIP) or generated via segmentation models with filtering (e.g., Object365, OpenImages). These masks are generally accurate, aside from minor noise near extremities.

\begin{table}
    \centering
    \scriptsize
    \renewcommand{\arraystretch}{1.2}
    \begin{tabular}{c|ccc}
        \toprule
        $\sigma_2$ & mIoU $\uparrow$ & mIoU-inv $\uparrow$ & L1 $\downarrow$ \\
        \midrule
        0  & 92.4 & 60.3 & 0.082 \\
        4  & \underline{93.0} & 62.7 & \underline{0.074} \\
        8  & \textbf{93.2} & \textbf{64.1} & \textbf{0.073} \\
        16 & 92.9 & \underline{63.3} & 0.076 \\
        32 & 92.9 & 62.6 & 0.075 \\
        \bottomrule
    \end{tabular}
    \vspace{1mm}
    \caption{Ablation study on the Gaussian standard deviation $\sigma_2$ used in occluded joint heatmap generation on AHP real dataset. The best values are shown in \textbf{bold}, and the second-best in \underline{underlined}.}
    \label{suppl_tab:c}
    \vspace{-4mm}
\end{table}

\begin{table}
    \centering
    \scriptsize
    \renewcommand{\arraystretch}{1.2}
    \begin{tabular}{c|ccc}
        \toprule
        $\sigma_1$ & mIoU $\uparrow$ & mIoU-inv $\uparrow$ & L1 $\downarrow$ \\
        \midrule
        1  & 92.9 & 62.2 & 0.078 \\
        2.5  & \textbf{93.2} & \textbf{64.1} & \textbf{0.073} \\
        5 & \underline{93.1} & \underline{63.3} & \underline{0.074} \\
        10 & 92.8 & 62.4 & 0.078 \\
        \bottomrule
    \end{tabular}
    \vspace{1mm}
    \caption{Ablation study on the Gaussian standard deviation $\sigma_1$ used in 2D joint heatmap generation.}
    \label{suppl_tab:d}
    \vspace{-4mm}
\end{table}

\begin{figure}
    \centering
    \includegraphics[width=\linewidth]{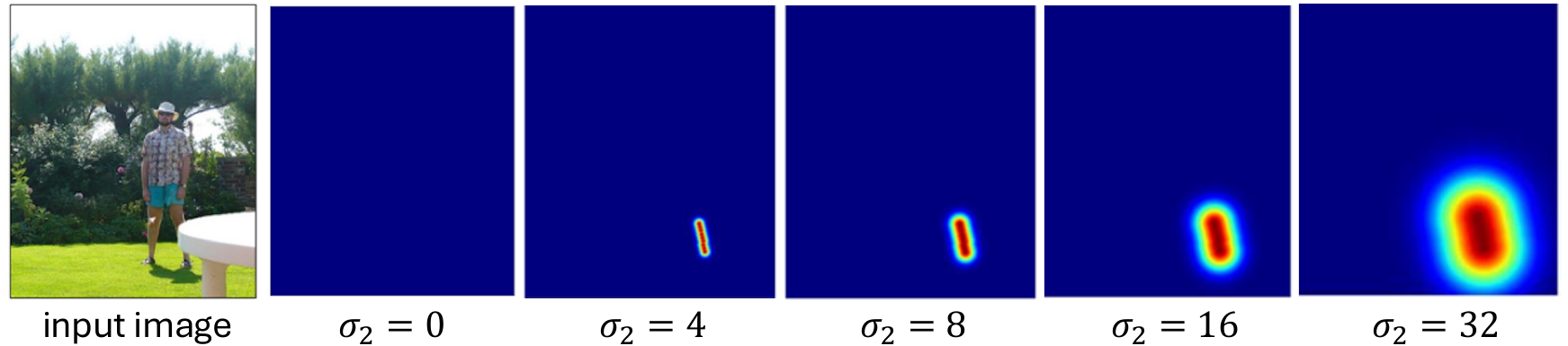}
    \vspace{-6mm}
    \caption{Visualization of occluded joint heatmaps generated with different values of $\sigma_2$.}
    \label{suppl_fig:b}
    \vspace{-4mm}
\end{figure}

\section{Additional Ablation Studies}

\subsection{Standard Deviation in Joint Heatmaps}
\label{suppl_subsec:b1}

We conduct an ablation study to investigate the effect of the Gaussian standard deviation $\sigma_2$ generating the occluded joint heatmap $H_\text{o}$. The parameter $\sigma_2$ controls the spatial spread of each joint in the heatmap. While the standard deviation for the 2D joint heatmap $H_\text{2D}$ is fixed at $\sigma_1=2.5$, following the configuration used in DPMesh~\cite{zhu2024dpmesh}, we vary $\sigma_2$ from 0 to 32 and evaluate the resulting performance on the mask completion task.

As shown in Table~\ref{suppl_tab:c}, the performance metrics---mIoU, mIoU-inv, and L1 loss---improve as $\sigma_2$ increases from 0, reaching optimal performance at $\sigma_2=8$. Beyond this point, further increases in $\sigma_2$ result in performance degradation. Based on these findings, we set $\sigma_2=8$ in our final configuration.

Excluding the case where the occluded joint heatmap $H_\text{o}$ is entirely omitted (i.e., $\sigma_2=0$), our method demonstrates low sensitivity to the choice of $\sigma_2$. Varying $\sigma_2$ between 0.5 and 4.0 times its final value (i.e., from 4 to 32) yields max changes of 0.32\% in mIoU, 2.34\% in mIoU-inv, and 4.11\% in L1. As shown in Table~\ref{suppl_tab:d}, a similar evaluation on $\sigma_1$ demonstrates comparable robustness, with maximum changes of 0.43\%, 2.96\%, and 6.85\% in mIoU, mIoU-inv, and L1, respectively. These results indicate that our method exhibits low sensitivity to both $\sigma_1$ and $\sigma_2$.

Fig.~\ref{suppl_fig:b} visualizes the occluded joint heatmaps $H_\text{o}$ generated using different values of $\sigma_2$. As $\sigma_2$ increases, the Gaussian activations become more spatially diffuse, covering larger areas centered on each joint. When $\sigma_2=8$, the heatmaps closely match the expected spatial extent of occluded body parts. However, excessively large values of $\sigma_2$ lead to over-spread activations that extend beyond the relevant occluded regions, reducing the localization precision and consequently degrading mask completion performance, as corroborated by the quantitative results in Table~\ref{suppl_tab:c}.

\begin{figure}
    \centering
    \includegraphics[width=\linewidth]{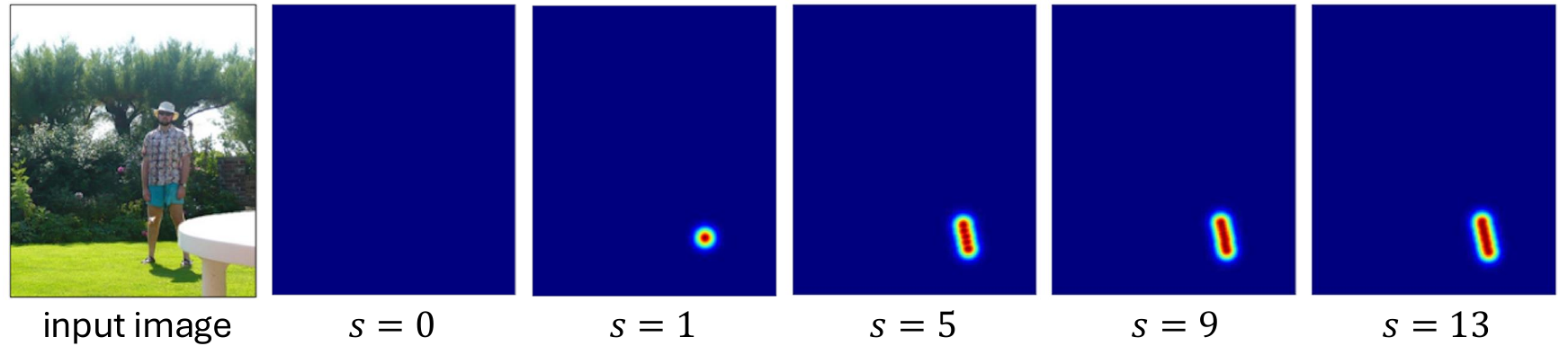}
    \vspace{-6mm}
    \caption{Visualization of occluded joint heatmaps generated with varying numbers of subdivided joints $s$.}
    \label{suppl_fig:c}
    \vspace{-4mm}
\end{figure}

\subsection{Number of Subdivided Joints}

An ablation study is conducted to assess the impact of the number of subdivided joints on the overall performance of the mask completion network. Based on the predefined joint pairs in Table~\ref{suppl_tab:a}, we vary the number of interpolated joints per pair to evaluate how the granularity of the resulting heatmaps influences mask quality. The Gaussian standard deviations $\sigma_1$ and $\sigma_2$ are fixed at 2.5 and 8, respectively, as determined in Section~\ref{suppl_subsec:b1}. Quantitative results are summarized in Table~\ref{suppl_tab:e}, and corresponding heatmaps are visualized in Fig.~\ref{suppl_fig:c}.

We define $s$ as the number of interpolated joints generated per joint pair. For example, $s=1$ corresponds to a single joint placed at the midpoint of each segment, while $s=3$ indicates three joints generated by dividing the segment into four equal intervals. Starting from $s=0$, we incrementally increase the value and observe a consistent improvement in mask completion performance. As shown in Table~\ref{suppl_tab:e}, the best performance is achieved at $s=9$. Beyond this point, performance begins to decline. Accordingly, we set $s=9$ in our final configuration, resulting in a total of 126 interpolated joints across 14 joint pairs (each divided into 10 segments).

As depicted in Fig.~\ref{suppl_fig:c}, the number of subdivided joints directly affects the structure of the occluded joint heatmap $H_\text{o}$. When $s$ is small, the heatmap lacks sufficient coverage of occluded body regions due to sparse joint placement. As $s$ increases, the joint heatmap becomes more detailed and accurately delineates occluded areas. However, setting $s$ beyond 9 leads to oversaturated and overly dense heatmaps, which do not yield further improvements in performance. These findings support the adoption of $s=9$ as the optimal setting.

\begin{table}
    \centering
    \scriptsize
    \renewcommand{\arraystretch}{1.2}
    \begin{tabular}{c|ccc}
        \toprule
        $s$ & mIoU $\uparrow$ & mIoU-inv $\uparrow$ & L1 $\downarrow$ \\
        \midrule
        0  & 92.8  & 62.2  & 0.076 \\
        1  & 92.9  & 62.7  & 0.076 \\
        3  & 93.1  & 63.2  & 0.074 \\
        5  & \textbf{93.4}  &  \underline{64.4}  & \underline{0.071} \\
        7  & \underline{93.2}  & 64.1  & 0.073 \\
        9  & \textbf{93.4}  & \textbf{64.9}  & \textbf{0.070} \\
        11 & 93.1  &  63.2  &  0.075 \\
        13 & 93.0  & 62.8  & 0.075 \\
        \bottomrule
    \end{tabular}
    \vspace{1mm}
    \caption{Ablation study on the number of subdivided joints $s$ per joint pairs.}
    \label{suppl_tab:e}
    \vspace{-4mm}
\end{table}

\begin{table}
    \centering
    \scriptsize
    \renewcommand{\arraystretch}{1.2}
    \begin{tabular}{l|ccc}
        \toprule
        Input Type & mIoU $\uparrow$ & mIoU-inv $\uparrow$ & L1 $\downarrow$ \\
        \midrule
        occluded joint mask & 93.0 & 62.9 & 0.074 \\
        whole joint heatmap & \underline{93.2} & \underline{63.9} & \underline{0.073} \\
        occluded joint heatmap & \textbf{93.4} & \textbf{64.9} & \textbf{0.070} \\
        \bottomrule
    \end{tabular}
    \vspace{1mm}
    \caption{Ablation study on different occlusion input variants.}
    \label{suppl_tab:f}
    \vspace{-4mm}
\end{table}

\subsection{Occlusion Input Variants}

We perform an ablation study to evaluate the impact of different input representations for occluded regions on mask completion performance. In particular, we compare three input types: the occluded joint mask, the whole joint heatmap, and the occluded joint heatmap.

The occluded joint mask is constructed following the same initial procedure used to generate the occluded joint heatmap, as detailed in Section~\ref{suppl_subsec:a1}---namely, extracting the occluded joints $J_\text{o}$. Instead of applying a Gaussian kernel, we construct a binary mask $M_\text{o}$ over the same spatial resolution as the heatmap. The final occluded joint mask is obtained by computing the intersection between $M_\text{o}$ and the complement of the modal mask, i.e., $(1 - M_\text{m})$, yielding a binary mask that highlights occluded regions not covered by $M_\text{m}$. 

The whole joint heatmap is generated by applying Gaussian kernels to all joints in the dense set $J_\text{dense}$, which includes both visible and occluded joints. Unlike the occluded joint heatmap, it does not require identifying occluded joints $J_\text{o}$ beforehand.

\begin{figure}
    \centering
    \includegraphics[width=0.9\linewidth]{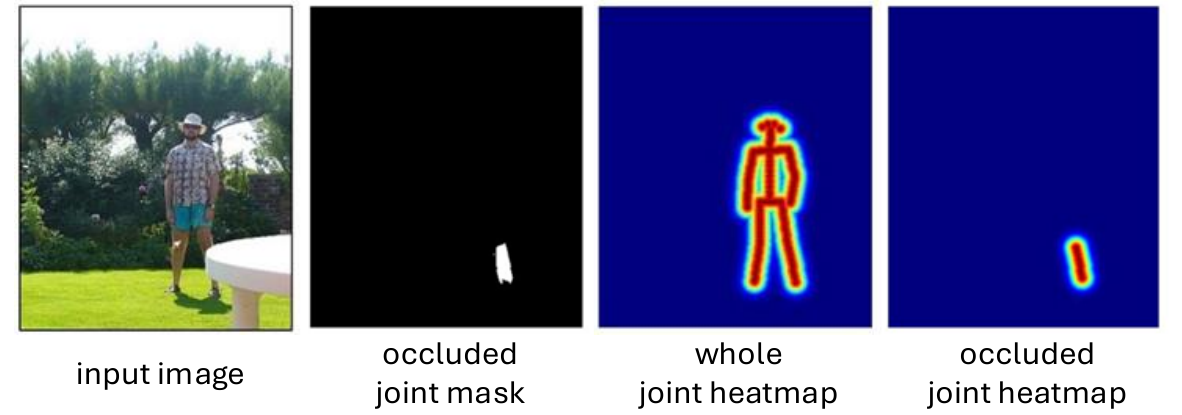}
    \vspace{-4mm}
    \caption{Visualization of different occlusion input variants.}
    \label{suppl_fig:d}
\end{figure}

Fig.~\ref{suppl_fig:d} presents visual comparisons of the different input representations, and Table~\ref{suppl_tab:f} reports the corresponding quantitative results. We observe that heatmap-based inputs consistently outperform the binary mask, suggesting that continuous-valued spatial cues are more informative for guiding mask completion. Among the evaluated representations, the occluded joint heatmap achieves the highest performance, as it explicitly encodes occluded areas that are not already represented in the modal mask $M_\text{m}$. In contrast, the whole joint heatmap introduces redundancy by including visible joints already captured by $M_\text{m}$, which diminishes its utility in isolating occluded regions.

\begin{table}
    \centering
    \scriptsize
    \renewcommand{\arraystretch}{1.2}
    \begin{tabular}{l|ccc|ccc}
        \toprule
        \multirow{2}{*}{Loss + Optimizer} & \multicolumn{3}{c|}{AHP real dataset} & \multicolumn{3}{c}{AHP syn dataset} \\
        \cline{2-4} \cline{5-7}
        & mIoU $\uparrow$ & mIoU-inv $\uparrow$ & L1 $\downarrow$
        & mIoU $\uparrow$ & mIoU-inv $\uparrow$ & L1 $\downarrow$ \\ 
        \midrule
        BCE + SGD & \textbf{93.5} & \textbf{65.1} & \textbf{0.070} & \textbf{92.6} & \textbf{67.1} & \textbf{0.105} \\
        CE + SGD & \underline{93.4} & \underline{64.9} & \textbf{0.070} & \underline{92.2} & \underline{66.0} & \underline{0.111} \\
        BCE + Adam & 91.9 & 59.7 & \underline{0.086} & 91.0 & 63.1 & 0.131 \\
        CE + Adam & 91.8 & 59.8 & 0.089 & 90.9 & 62.6 & 0.133 \\
        \bottomrule
    \end{tabular}
    \vspace{1mm}
    \caption{Ablation study on different optimizer and loss function configurations for mask completion.}
    \label{suppl_tab:g}
    \vspace{-4mm}
\end{table}

\subsection{Loss Configuration for Mask Completion}

We evaluate various loss functions and optimization strategies to determine their impact on mask completion performance. Specifically, we consider four configurations: BCE with stochastic gradient descent with momentum (SGD), cross-entropy (CE) with SGD, BCE with Adam, and CE with Adam. As summarized in Table~\ref{suppl_tab:g}, SGD consistently outperforms Adam across all evaluation metrics. Furthermore, BCE demonstrates more stable generalization compared to CE, particularly on the synthetic dataset. While BCE and CE yield comparable results on real-world data for both optimizers, BCE provides superior performance on synthetic data. Based on these findings, we adopt BCE loss with the SGD optimizer for the mask completion network in our proposed approach.

\begin{table}
    \centering
    \scriptsize
    \renewcommand{\arraystretch}{1.2}
    \begin{tabular}{l|cccccc}
        \toprule
        Step & L1* $\downarrow$ & FID $\downarrow$ & LPIPS* $\downarrow$ & PSNR $\uparrow$ & MSE* $\downarrow$ & Time \\
        \midrule
        0 & 9.58 & 9.86 & 47.27 & 26.27 & 2.98 & 0s \\
        10 & 6.99 & 14.30 & 49.39 & 29.02 & \underline{1.89} & 1s \\
        30 & 6.29 & 9.75 & 40.94 & \textbf{29.33} & \textbf{1.84} & 3s \\
        50 & \textbf{6.15} & 8.26 & 37.63 & \underline{29.13} & 1.91 & 5s \\
        70 & \textbf{6.15} & 7.94 & 36.07 & 28.92 & 1.96 & 7s \\
        100 & \underline{6.16} & 7.50 & 35.08 & 28.76 & 2.00 & 10s \\
        150 & 6.18 & \textbf{7.20} & \underline{34.43} & 28.63 & 2.04 & 15s \\
        200 & 6.20 & \underline{7.46} & \textbf{34.12} & 28.55 & 2.06 & 20s \\
        \bottomrule
        \end{tabular}
    \vspace{1mm}
    \caption{Quantitative results for the whole image. L1*, LPIPS*, and MSE* values are scaled by a factor of 1000 for readability.}
    \label{suppl_tab:h}
    \vspace{-4mm}
\end{table}

\begin{table}
    \centering
    \scriptsize
    \renewcommand{\arraystretch}{1.2}
    \begin{tabular}{l|ccccc}
        \toprule
        Step & L1* $\downarrow$ & FID $\downarrow$ & LPIPS* $\downarrow$ & PSNR $\uparrow$ & MSE* $\downarrow$ \\
        \midrule
        0 & 3.28 & \textbf{8.98} & 14.13 & 30.39 & 1.73 \\
        10 & \textbf{2.86} & \underline{9.54} & 13.98 & \textbf{32.36} & \textbf{1.37} \\
        30 & \underline{2.98} & 10.39 & \underline{13.65} & \underline{31.77} & \underline{1.49} \\
        50 & 3.10 & 10.48 & 13.68 & 31.25 & 1.58 \\
        70 & 3.16 & 11.21 & 13.73 & 30.96 & 1.63 \\
        100 & 3.18 & 11.43 & 13.70 & 30.81 & 1.66 \\
        150 & 3.19 & 10.62 & \textbf{13.64} & 30.70 & 1.68 \\
        200 & 3.21 & 10.02 & 13.68 & 30.62 & 1.69 \\
        \bottomrule
    \end{tabular}
    \vspace{1mm}
    \caption{Quantitative results for the invisible region.}
    \label{suppl_tab:i}
    \vspace{-4mm}
\end{table}

\begin{table}
    \centering
    \scriptsize
    \renewcommand{\arraystretch}{1.2}
    \begin{tabular}{l|ccccc}
        \toprule
        Step & L1* $\downarrow$ & FID $\downarrow$ & LPIPS* $\downarrow$ & PSNR $\uparrow$ & MSE* $\downarrow$ \\
        \midrule
        0 & 6.29 & 7.16 & 38.09 & 29.69 & 1.24 \\
        10 & 4.13 & 10.00 & 39.67 & 33.80 & 0.52 \\
        30 & 3.30 & 4.53 & 31.02 & 35.56 & 0.36 \\
        50 & 3.06 & 4.24 & 27.51 & \textbf{35.91} & \textbf{0.33} \\
        70 & \underline{2.99} & 4.36 & 25.96 & \underline{35.78} & \textbf{0.33} \\
        100 & \textbf{2.98} & 4.00 & 25.07 & 35.50 & \underline{0.34} \\
        150 & \textbf{2.98} & \textbf{3.38} & \underline{24.49} & 35.23 & 0.36 \\
        200 & \underline{2.99} & \underline{3.53} & \textbf{24.18} & 35.09 & 0.37 \\
        \bottomrule
    \end{tabular}
    \vspace{1mm}
    \caption{Quantitative results for the visible region.}
    \label{suppl_tab:j}
    \vspace{-4mm}
\end{table}

\begin{figure*}
    \centering
    \includegraphics[width=0.9\linewidth]{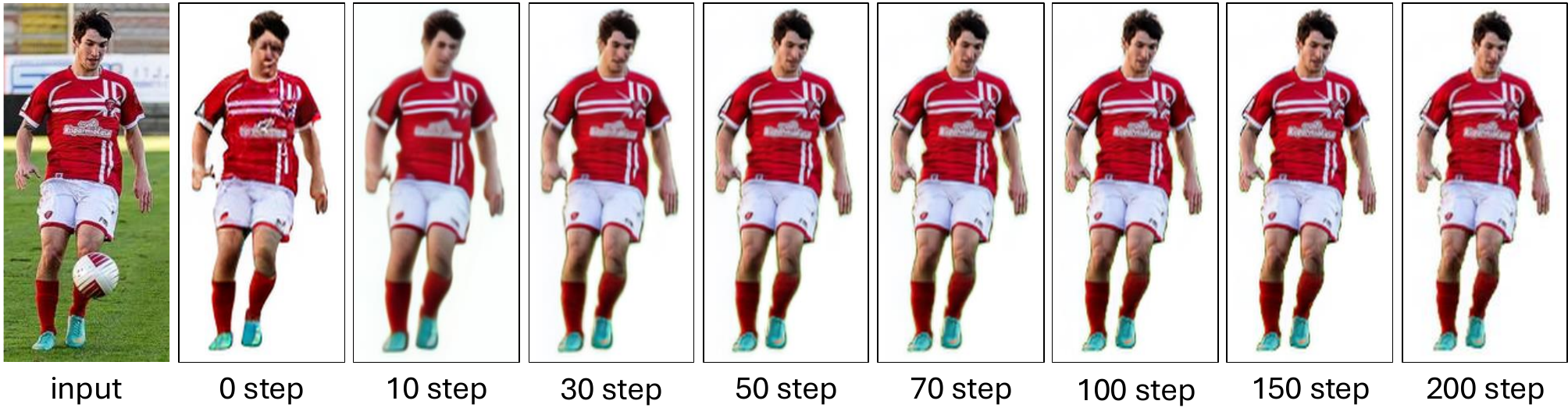}
    \vspace{-4mm}
    \caption{Effect of optimization steps on decoder fine-tuning results.}
    \label{suppl_fig:e}
    \vspace{-2mm}
\end{figure*}

\subsection{Decoder Fine-tuning}

Decoder fine-tuning is performed at inference time, with the number of optimization steps treated as a controllable hyperparameter. To determine an appropriate value, we conduct an ablation study by varying the number of steps from 0 to 200 and evaluating their impact on reconstruction quality. Quantitative results are reported in Tables~\ref{suppl_tab:h},~\ref{suppl_tab:i}, and~\ref{suppl_tab:j}, corresponding to the whole image, the invisible region, and the visible region, respectively.

As shown in Table~\ref{suppl_tab:i}, applying decoder fine-tuning with a small number of steps (0 to 10) improves performance in the invisible region. This improvement stems from the fact that the initial output at step 0, as illustrated in Fig.~\ref{suppl_fig:e}, often exhibits global degradation. Early fine-tuning helps restore overall image quality, thereby indirectly enhancing the occluded areas. Beyond this range, performance gradually declines as the number of steps increases, particularly in terms of L1, FID, PSNR, and MSE. This trend aligns with the purpose of decoder fine-tuning: to refine the visible region to more closely match the input image, potentially at the expense of accuracy in the occluded regions. Additionally, as shown in Table~\ref{suppl_tab:h}, the total optimization time increases approximately linearly with the number of optimization steps.

In contrast, Table~\ref{suppl_tab:j} shows that in the visible region, L1, FID, and LPIPS metrics consistently improve as the number of optimization steps increases. During the early steps, image-space reconstruction metrics such as L1 and MSE improve rapidly, while perceptual scores like FID and LPIPS may temporarily degrade. However, these scores naturally improve as optimization continues (steps 10 to 30), leading to gradual overall improvements. For PSNR and MSE, performance peaks around 50 steps and gradually declines thereafter. These findings highlight a trade-off: increasing the number of optimization steps enhances fidelity in the visible region, but at the cost of reduced performance in the invisible region and increased computational overhead.

Based on these findings, we set the number of optimization steps to 50 in our final configuration. This choice offers a balanced trade-off, achieving peak performance in PSNR and MSE for the visible region. As illustrated in Fig.~\ref{suppl_fig:e}, most qualitative improvements occur within the early steps, with only minor changes observed beyond 50 steps.

\begin{figure}
    \centering
    \includegraphics[width=\linewidth]{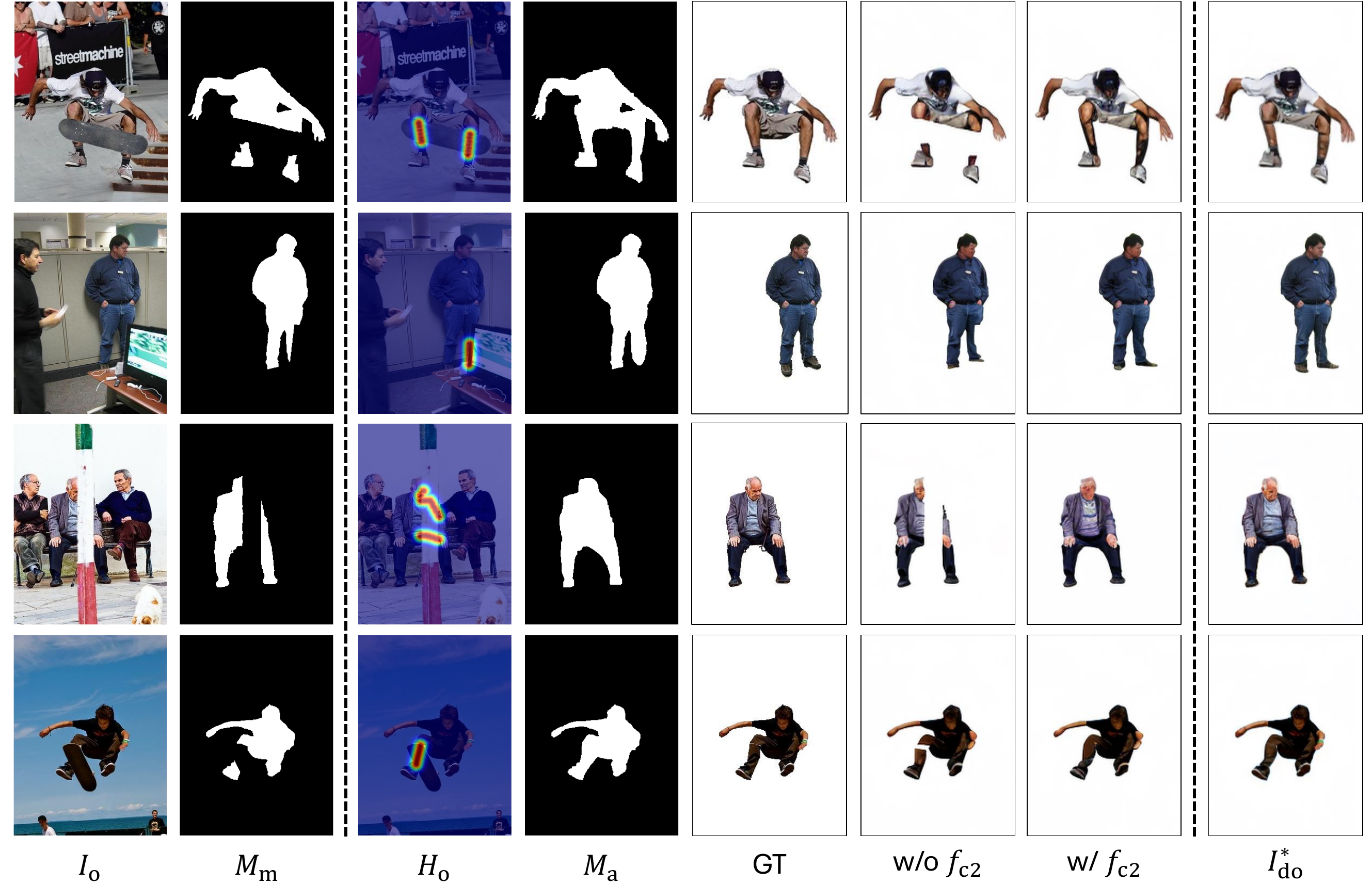}
    \vspace{-7mm}
    \caption{Effect of control feature $f_\text{c2}$ on RGB completion.}
    \label{suppl_fig:f}
\end{figure}

\subsection{Control Feature for RGB Completion}

The mask completion network predicts the amodal mask $M_\text{a}$ by utilizing the occluded joint heatmap $H_\text{o}$ and a diffusion-based human body prior. This amodal mask is then processed by ControlNet to generate the control feature $f_\text{c2}$, which provides spatial guidance to the RGB completion network by indicating the regions that require reconstruction.

As shown in Fig.~\ref{suppl_fig:f}, excluding $f_\text{c2}$ during RGB completion often results in inaccurate localization of occluded areas, leading to incomplete reconstruction. In contrast, incorporating $f_\text{c2}$ enables the network to more effectively identify where reconstruction is needed, thereby facilitating improved RGB completion. This highlights the importance of $f_\text{c2}$ in explicitly guiding the network toward the intended reconstruction regions. The final de-occluded image $I_\text{do}^*$, generated using $f_\text{c2}$ in conjunction with human-specific text features and decoder fine-tuning, shows improved RGB reconstruction quality and overall perceptual fidelity.

\subsection{Text feature for RGB Completion}
\label{suppl_subsec:b7}

In Table 5 and Fig. 6 of the main paper, we show that our proposed human-specific text feature $f_\text{hs}$, derived from a VQA-based approach, outperforms CLIP-based features extracted directly from the input image $I_\text{o}$. However, since this comparison involves CLIP features extracted from an image and text features derived from a VQA-based prompt, it does not directly quantify the contribution of the VQA-based prompt itself.

To address this, we conduct an additional ablation study comparing our method against a generic text prompt (GTP)-based alternative, where both methods use text-only inputs. Following ~\cite{zhao2023unleashing}, we adopt the GTP “a photo of a human” to obtain CLIP-based text features. In contrast, our VQA-based approach generates high-level, human-specific semantic prompts such as “a person wearing a red shirt and jeans,” which provide more detailed appearance cues.

As shown in Table~\ref{suppl_tab:k}, our method achieves better performance across all metrics (L1, FID, LPIPS, PSNR) compared to the GTP-based variant. These results show the benefit of VQA-based, human-centric descriptions over generic prompts.

\begin{table}
    \centering
    \scriptsize
    \renewcommand{\arraystretch}{1.2}
    \begin{tabular}{l|cccc}
        \toprule
        Method & L1* $\downarrow$ & FID $\downarrow$ & LPIPS* $\downarrow$ & PSNR $\uparrow$ \\
        \midrule
        Ours w/ GTP & 7.67 & 6.61 & 29.14 & 27.99 \\
        Ours w/ VQA & \textbf{7.18} & \textbf{5.46} & \textbf{27.86} & \textbf{28.36} \\
        \bottomrule
    \end{tabular}
    \vspace{1mm}
    \caption{Comparison of VQA-based and generic text prompts.}
    \label{suppl_tab:k}
    \vspace{-4mm}
\end{table}

\begin{table}
    \centering
    \scriptsize
    \renewcommand{\arraystretch}{1.2}
    \begin{tabular}{l|ccc}
        \toprule
        Training Setting & mIoU $\uparrow$ & mIoU-inv $\uparrow$ & L1 $\downarrow$ \\
        \midrule
        Ours w/ only AM & 90.2 & 54.4 & 0.109 \\
        Ours W/ AM \& IAM & \textbf{92.1} & \textbf{61.5} & \textbf{0.086} \\
        \bottomrule
    \end{tabular}
    \vspace{1mm}
    \caption{Ablation study on inaccurate input masks for mask completion.}
    \label{suppl_tab:l}
    \vspace{-4mm}
\end{table}

\subsection{Inaccurate Input Masks}
\label{suppl_subsec:b8}

Following prior works~\cite{liu2024object,zhan2020self}, our method assumes error-free input modal masks $M_\text{m}$. However, acquiring accurate masks in real-world scenarios can be challenging. To assess the robustness of our framework when provided with inaccurate mask inputs, we conduct an ablation study using noisy masks.

We consider two representative types of mask errors: (1) incomplete masks missing parts of the person, or (2) masks including background or other objects. Our method handles the first type well by treating missing regions as occlusions. However, the second type may reduce performance.

To address this, we apply a simple data augmentation strategy during training: injecting inaccurate masks with a probability of 10\%. These inaccurate masks (IAM) are generated by randomly expanding the original mask boundary to include an additional 5\%–30\% area. Evaluation on test images with similar types of mask errors shows improved robustness. As shown in Table~\ref{suppl_tab:l}, training with both accurate masks (AM) and inaccurate masks (IAM) outperforms training with only AM in terms of mIoU, mIoU-inv, and L1 loss.

\begin{table}
    \centering
    {\scriptsize
    \renewcommand{\arraystretch}{1.2}
    \begin{tabular}{l|ccccc}
        \toprule
        Method & P2S $\downarrow$ & Chamfer $\downarrow$ & Normal $\downarrow$ & MSE $\downarrow$ & LPIPS $\downarrow$ \\
        \midrule
        PIFu w/ $I_\text{o}$ & 21.54 & 22.40 & 0.523 & 0.401 & 0.325 \\
        PIFu w/ $I_\text{do}^*$ & \textbf{10.95} & \textbf{10.81} & \textbf{0.318} & \textbf{0.247} & \textbf{0.216} \\
        \midrule
        PIFu w/ GT & 9.32 & 9.36 & 0.276 & 0.203 & 0.198 \\
        \bottomrule
    \end{tabular}
    }
    \vspace{1mm}
    \caption{Quantitative comparison on 3D clothed human reconstruction.}
    \vspace{-4mm}
    \label{suppl_tab:m}
\end{table}

\subsection{Effect on Downstream Task}
\label{suppl_subsec:b9}

To evaluate the utility of our de-occluded image $I_\text{do}^*$ in downstream applications, we assess its effectiveness on a representative task: 3D clothed human reconstruction. Specifically, we employ the PIFu model~\cite{saito2019pifu} to reconstruct 3D meshes from images and conduct experiments on the THuman2.0 dataset~\cite{yu2021function4d}, which provides GT 3D meshes. Following the procedure described in Section~\ref{suppl_subsec:a5}, we create 40 test samples by applying synthetic occlusions to images from the THuman2.0 dataset instead of AHP.

We compare three configurations to assess the impact of de-occlusion: (1) PIFu w/ $I_\text{o}$, which takes the occluded image as input; (2) PIFu w/ $I_\text{do}^*$, which uses the de-occluded image produced by our method; and (3) PIFu w/ GT, which serves as an upper bound by using the non-occluded GT image. All configurations employ the same pretrained PIFu model without any additional fine-tuning.

We evaluate both geometric accuracy and texture quality using five standard metrics: point-to-surface (P2S) distance, Chamfer distance, and surface normal error for geometry; and MSE and LPIPS for texture. As shown in Table~\ref{suppl_tab:m}, using our de-occluded image $I_\text{do}^*$ as input results in substantial improvements across all metrics compared to using the occluded image $I_\text{o}$. Notably, PIFu w/ $I_\text{do}^*$ achieves performance that closely approaches PIFu w/ GT, indicating that our human de-occlusion network can effectively generate de-occluded images that closely resemble the GT, despite not being trained on the THuman2.0 dataset.

\section{Additional Results}

\subsection{Mask Completion}

\begin{figure}
    \centering
    \includegraphics[width=\linewidth]{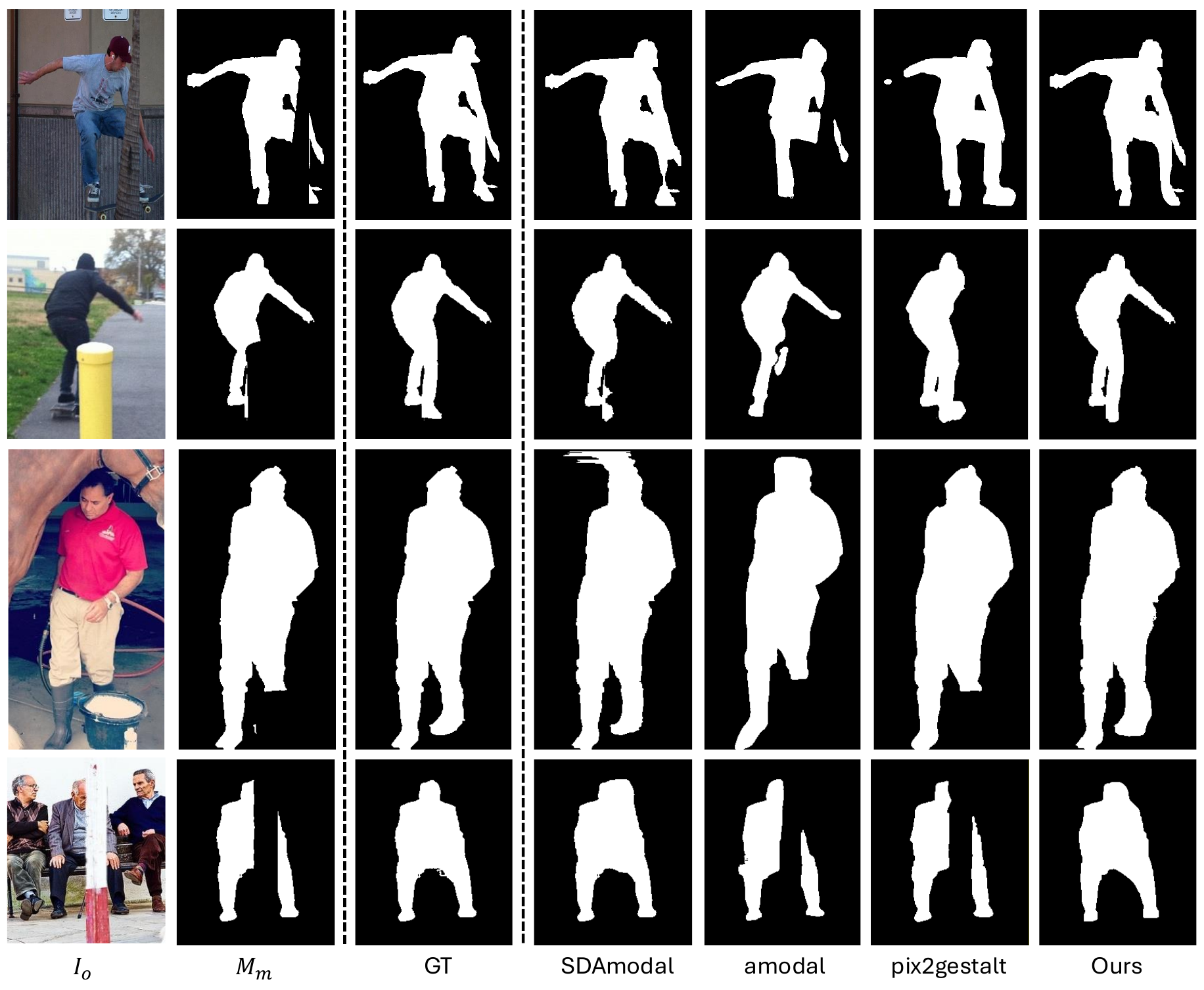}
    \vspace{-7mm}
    \caption{Additional qualitative comparison of mask completion results.}
    \label{suppl_fig:g}
    \vspace{-4mm}
\end{figure}

We present additional qualitative results of the mask completion network in Fig.~\ref{suppl_fig:g}. Compared to other baselines, our method consistently achieves more accurate reconstruction of occluded regions. Specifically, \emph{SDAmodal} often fails to recover the missing parts, as shown in the first and second rows of Fig.~\ref{suppl_fig:g}, or generates excessive completions that extend beyond the body boundary, as seen in the third and fourth rows. \emph{amodal} struggles to infer the missing regions across all examples. While \emph{pix2gestalt} occasionally succeeds in recovering occluded areas, it frequently loses visible details (e.g., in the first and second rows) or fails entirely, similar to other methods in the third and fourth rows. In contrast, our mask completion network leverages a human body prior to accurately reconstruct occluded regions, while avoiding unrealistic over-completions that extend beyond the expected body structure.

\begin{figure}
    \centering
    \includegraphics[width=\linewidth]{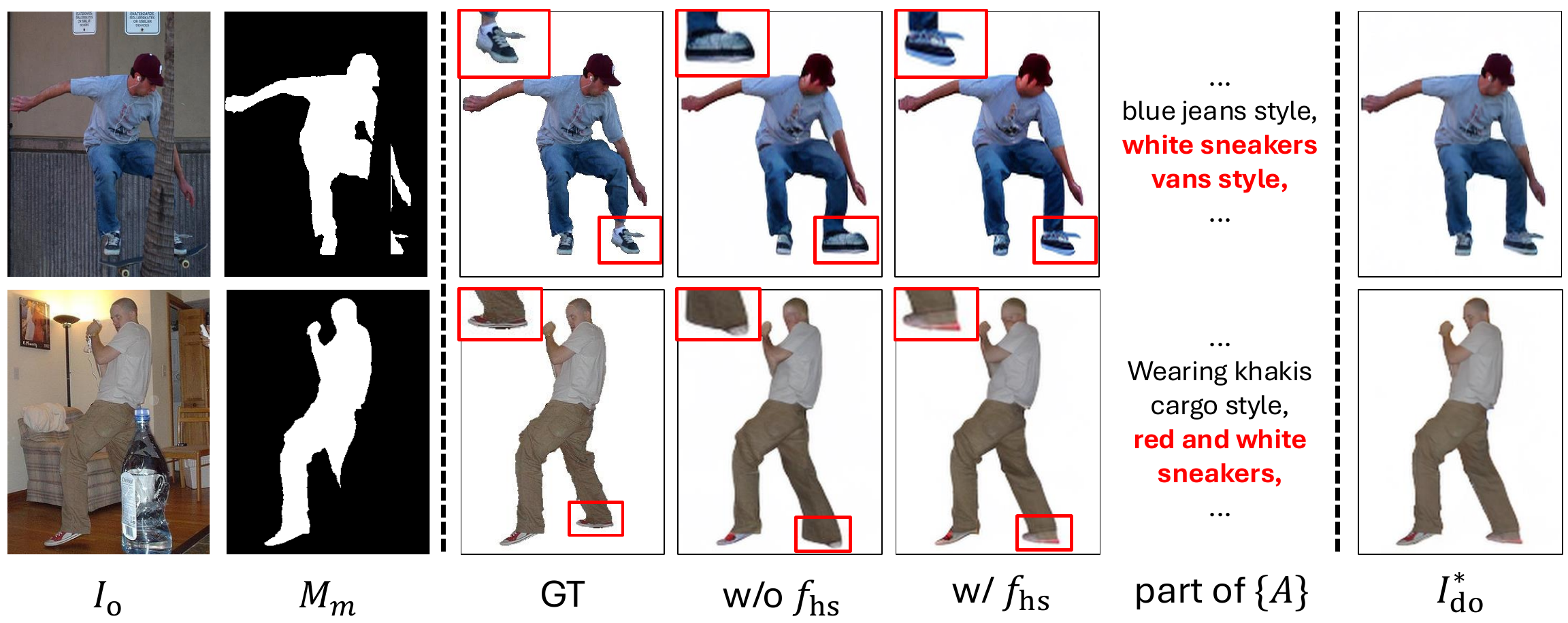}
    \vspace{-7mm}
    \caption{Additional qualitative results demonstrating the effect of the human-specific text feature $f_\text{hs}$ on RGB completion.}
    \label{suppl_fig:h}
    \vspace{-4mm}
\end{figure}

\subsection{Human-specific Text Feature}

Fig.~\ref{suppl_fig:h} presents extended qualitative results illustrating the impact of the human-specific text feature $f_\text{hs}$ on RGB completion. Without $f_\text{hs}$, the model often fails to preserve appearance details of visible regions or to reconstruct occluded parts. For example, in the first row of Fig.~\ref{suppl_fig:h}, the model loses fine-grained texture on the visible shoes. In the second row, it completely fails to reconstruct the occluded shoes. In contrast, when $f_\text{hs}$ is used---providing cues such as “Vans-style shoes” (first row) or “red and white sneakers” (second row)---the model successfully recovers appearances that closely resemble the GT. Notably, in the second row, the model infers the the occluded region based on the visible portion, guided by the textual information embedded in $f_\text{hs}$.

\begin{figure*}
    \centering
    \includegraphics[width=0.8\linewidth]{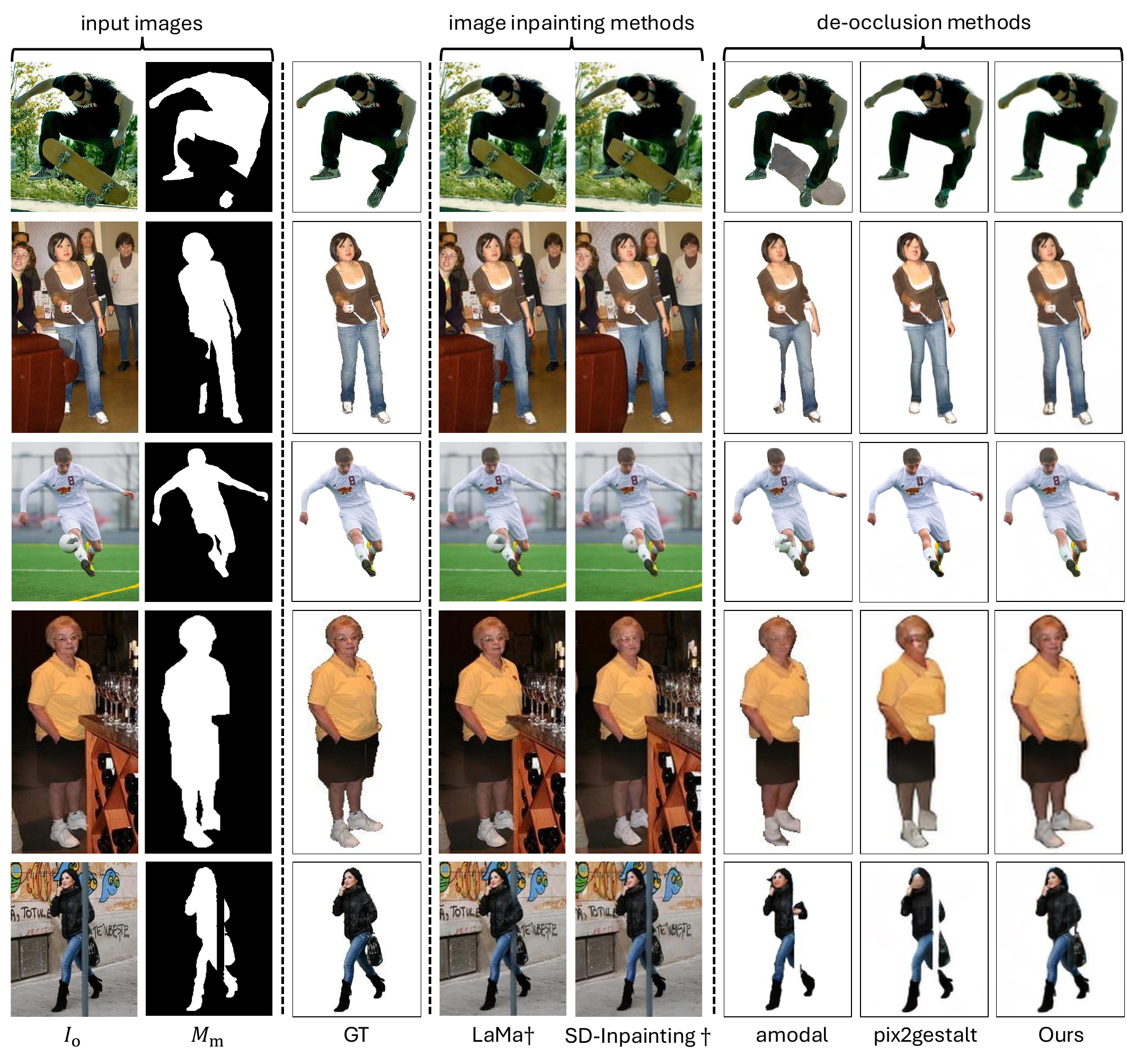}
    \vspace{-5mm}
    \caption{Additional qualitative results of RGB completion, including comparisons with image inpainting methods. \dag\ denotes the use of the GT invisible mask as input.}
    \label{suppl_fig:i}
\end{figure*}

\begin{figure}
    \centering
    \includegraphics[width=\linewidth]{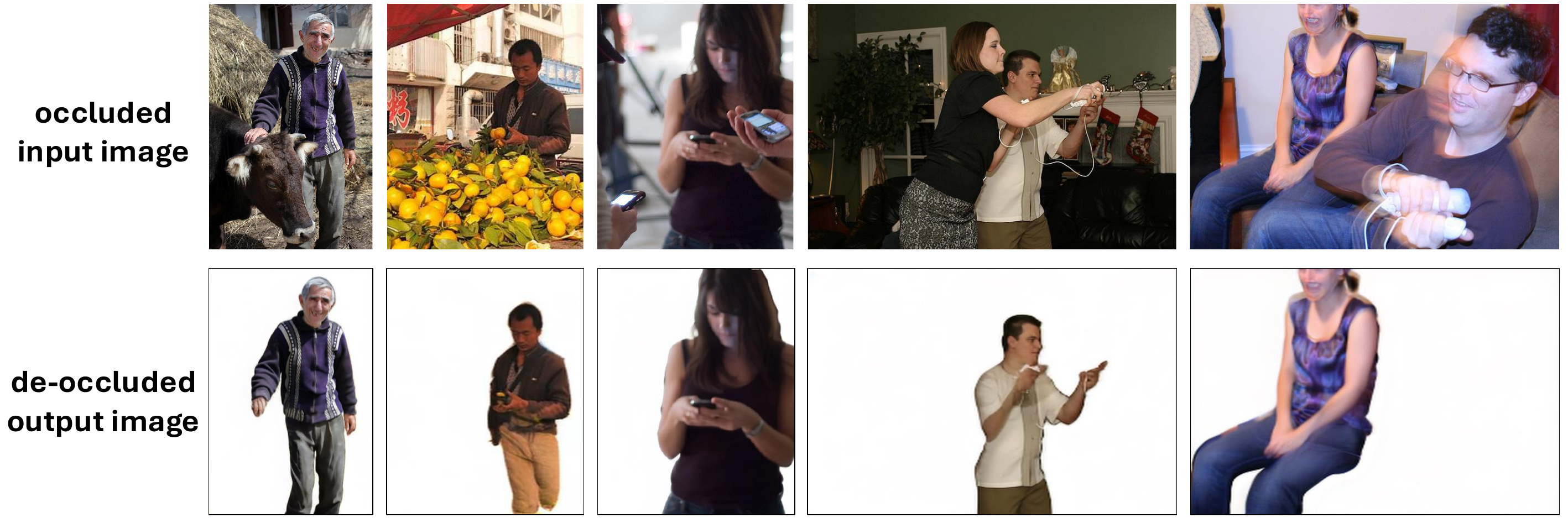}
    \vspace{-7mm}
    \caption{Qualitative results of RGB completion on in-the-wild images from the COCOA dataset.}
    \label{suppl_fig:j}
\end{figure}

\subsection{RGB Completion}

Fig.~\ref{suppl_fig:i} presents additional qualitative results for RGB completion. In addition to de-occlusion methods, we compare our approach with two general-purpose image inpainting methods: \emph{Large Mask Inpainting} (LaMa)~\cite{suvorov2021lama} and \emph{Stable Diffusion Inpainting} (SD-Inpainting). These inpainting methods take the GT invisible mask as input, allowing them to perform inpainting specifically within the occluded region.

In contrast, de-occlusion methods do not rely on the GT invisible mask and must instead infer which regions require reconstruction. Despite having access to the true occluded regions, inpainting methods still fail to produce effective de-occlusion, as discussed in Section~2.2 of the main paper. Specifically, LaMa often produces outputs that are nearly identical to the input image, failing to reconstruct the occluded areas. SD-Inpainting, which operates in the latent space, not only fails to recover missing content but also tends to distort the visible regions. While LaMa---based on convolutional neural networks and not involving latent representations---preserves visible regions more reliably, it still struggles to reconstruct the occluded content effectively.

These results highlight the limitations of applying general-purpose inpainting techniques to the task of human de-occlusion. Among the de-occlusion baselines, \emph{amodal} often fails to localize occluded regions and frequently includes occluding objects in its output. \emph{pix2gestalt} demonstrates slightly improved localization but lacks sufficient human body priors for consistent reconstruction. Both methods also show noticeable degradation in visible appearance details, such as facial features and garment textures.

In contrast, our proposed method accurately identifies occluded regions and generates high-quality RGB completions. It not only reconstructs the missing content with higher fidelity but also preserves the visual integrity of the visible areas---addressing a major shortcoming of existing approaches that often degrade these regions during the completion process.

\begin{table}
    \centering
    \scriptsize
    \renewcommand{\arraystretch}{1.2}
    \begin{tabular}{l|ccc}
        \toprule
        Network & mIoU $\uparrow$ & mIoU-inv $\uparrow$ & L1 $\downarrow$ \\
        \midrule
        pix2gestalt~\cite{ozguroglu2024pix2gestalt} & 82.5 & 57.1 & 0.448 \\
        SDAmodal~\cite{zhan2024amodal} & 81.0 & 61.6 & 0.412 \\
        Ours & \textbf{83.4} & \textbf{64.1} & \textbf{0.347} \\
        \bottomrule
    \end{tabular}
    \vspace{1mm}
    \caption{Ablation study on mask completion in an in-the-wild setting.}
    \label{suppl_tab:n}
    \vspace{-6mm}
\end{table}

\subsection{Evaluation on In-The-Wild Occlusion}

Table 1 in the main paper presents quantitative results on the AHP dataset, demonstrating that the proposed method outperforms existing methods in mask completion. To further assess its effectiveness under less constrained conditions, we perform an additional quantitative evaluation on the COCOA dataset, which includes in-the-wild occlusions. We randomly sample 40 human instances from the COCOA validation set and evaluate mask completion performance using manually annotated GT masks. As the dataset does not provide corresponding GT RGB images, this evaluation is limited to the mask domain.

Table~\ref{suppl_tab:n} summarizes the quantitative results, where the proposed method consistently outperforms existing approaches across all evaluation metrics (mIoU, mIoU-inv, and L1) on this in-the-wild dataset.

Fig.~\ref{suppl_fig:j} presents RGB completion results on the same COCOA samples. While GT RGB images are unavailable, the qualitative examples suggest that the proposed method generates plausible completions under challenging scenarios, including multi-person interactions (fourth column) and motion blur (fifth column).

\subsection{Computational Complexity}

We report the inference time and memory usage of each stage in our framework. All measurements were conducted on a single NVIDIA RTX 3090 GPU.

In the mask completion stage, our method requires approximately 0.1 seconds per image, with peak GPU memory usage of around 14 GB. Compared to \emph{SDAmodal}, which takes about 2 seconds and consumes 17 GB, our approach is relatively lightweight in both runtime and memory usage.

In the RGB completion stage, inference takes around 22 seconds per image, with a memory usage of approximately 14 GB. The slower RGB process is due to the iterative denoising steps required by the DM. For comparison, \emph{pix2gestalt} takes about 20 seconds and 19 GB, and \emph{amodal} takes approximately 15 seconds and 16 GB.

\begin{figure}
    \centering
    \includegraphics[width=0.8\linewidth]{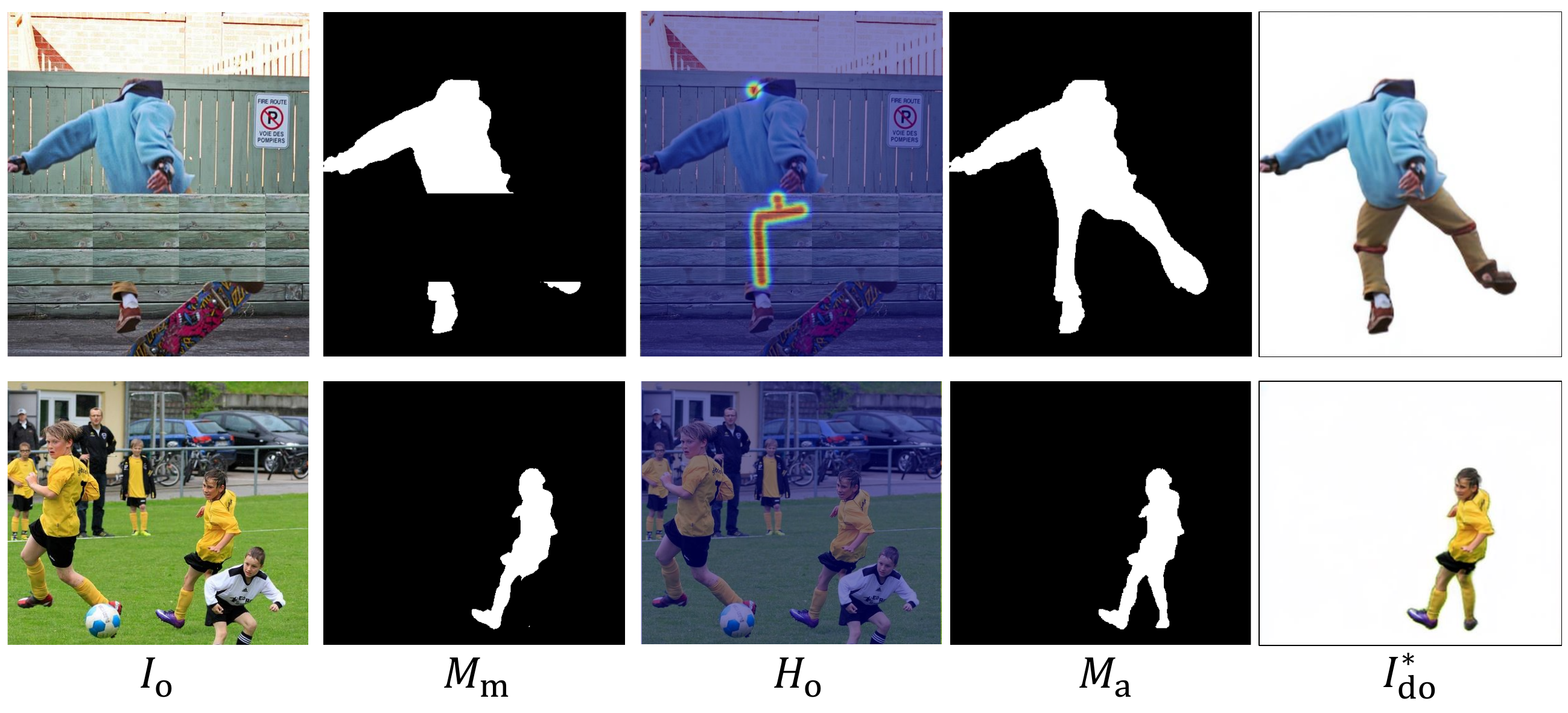}
    \vspace{-4mm}
    \caption{Qualitative examples of mask and RGB completion under poor 2D joint detection.}
    \label{suppl_fig:k}
    \vspace{-4mm}
\end{figure}

\subsection{Robustness to Inaccurate Pose Estimation}

Our framework utilizes the predicted 2D pose $J_\text{2D}$ from an off-the-shelf 2D pose estimator~\cite{8765346}, along with the derived occluded joint heatmap $H_\text{o}$, to guide the mask completion process. The resulting amodal mask $M_\text{a}$ is then used as input to the RGB completion module, which generates the final de-occluded image $I_\text{do}^*$.

While $J_\text{2D}$ generally provides a reliable human body prior under moderate occlusion, it may fail under severe occlusion. In our training and test datasets, approximately 7.9\% of samples exhibit poor 2D joint detection, defined as cases where six or fewer confident joints are detected out of 25 total keypoints.

These failure cases are included during training, enabling our model to learn robustness against inaccurate pose inputs. This strategy aligns with the approach described in Section~\ref{suppl_subsec:b8}, where the inclusion of inaccurate input masks during training improves generalization. As shown in Fig.~\ref{suppl_fig:k}, the top row presents an example where $H_\text{o}$ partially captures the occlusion, whereas the bottom row shows a case where $H_\text{o}$ entirely fails to represent the occlusion. Despite these limitations, our framework successfully completes the amodal mask $M_\text{a}$ and produces plausible de-occluded images $I_\text{do}^*$ in both cases.

\begin{figure}
    \centering
    \includegraphics[width=0.7\linewidth]{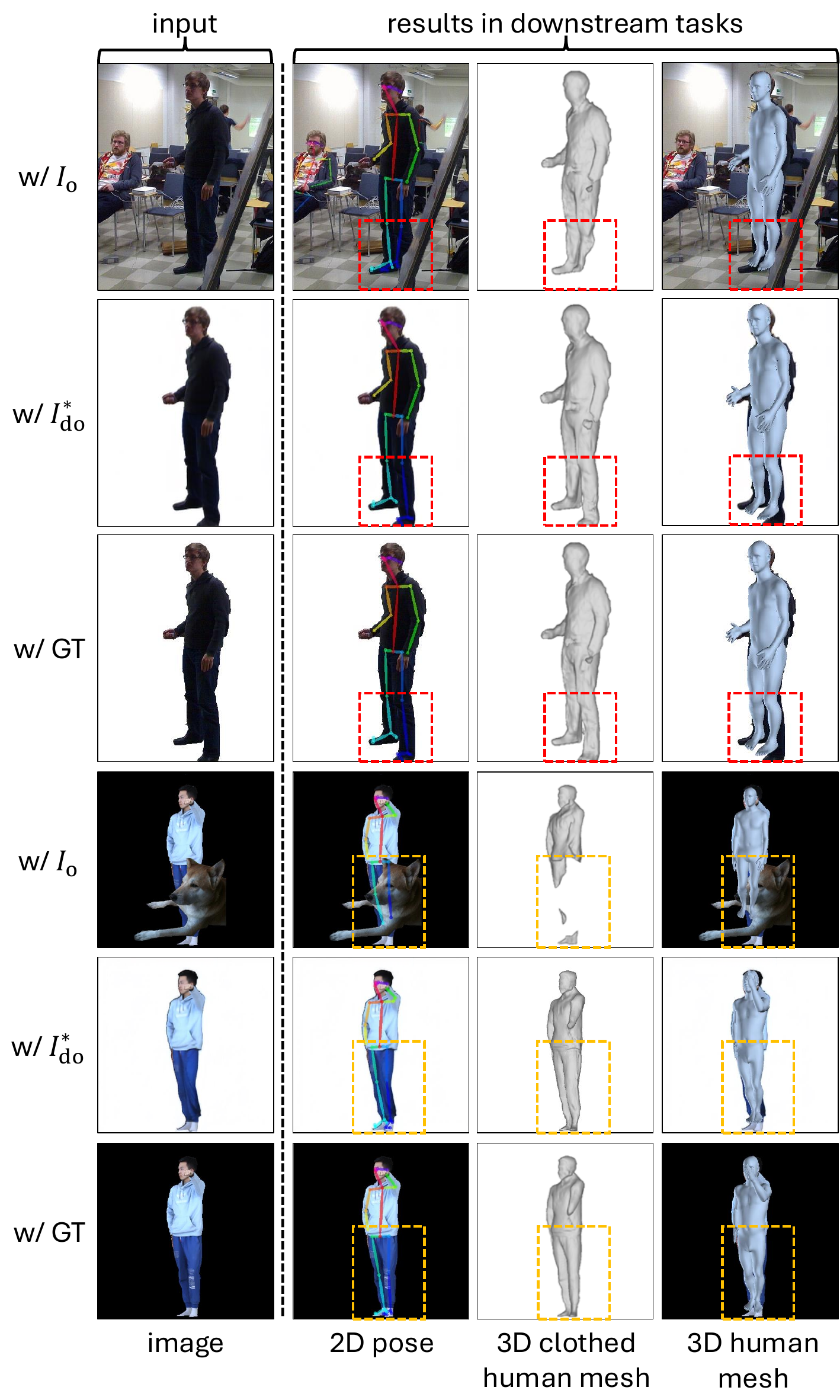}
    \vspace{-4mm}
    \caption{Comparison of downstream task results using occluded, de-occluded, and GT images. The top three rows present results on the AHP real dataset, while the bottom three rows show results on the occluded THuman2.0 dataset, as described in Section~\ref{suppl_subsec:b9}.}
    \vspace{-4mm}
    \label{suppl_fig:l}
\end{figure}

\subsection{Downstream Tasks}
\label{suppl_subsec:c7}

To evaluate the practical utility of our de-occluded image $I_\text{do}^*$, we assess its performance on several downstream tasks, including 2D pose estimation, 3D clothed human reconstruction, and 3D human mesh recovery. As illustrated in Fig.~\ref{suppl_fig:l}, we compare the results of these tasks using three types of input images: the original occluded image $I_\text{o}$, the de-occluded image $I_\text{do}^*$, and the GT image. The same downstream models are applied across all input types to ensure a fair comparison.

The results show that using $I_\text{do}^*$ yields more accurate pose and mesh predictions than using $I_\text{o}$. In particular, body parts that are either missing or inaccurately inferred in $I_\text{o}$ are better recovered when $I_\text{do}^*$ is used, leading to more complete and reliable reconstructions. Furthermore, the outputs produced from $I_\text{do}^*$ are visually comparable to those obtained from GT images, demonstrating the effectiveness of our de-occlusion framework in enabling high-level vision applications.

\begin{figure}
    \centering
    \includegraphics[width=0.9\linewidth]{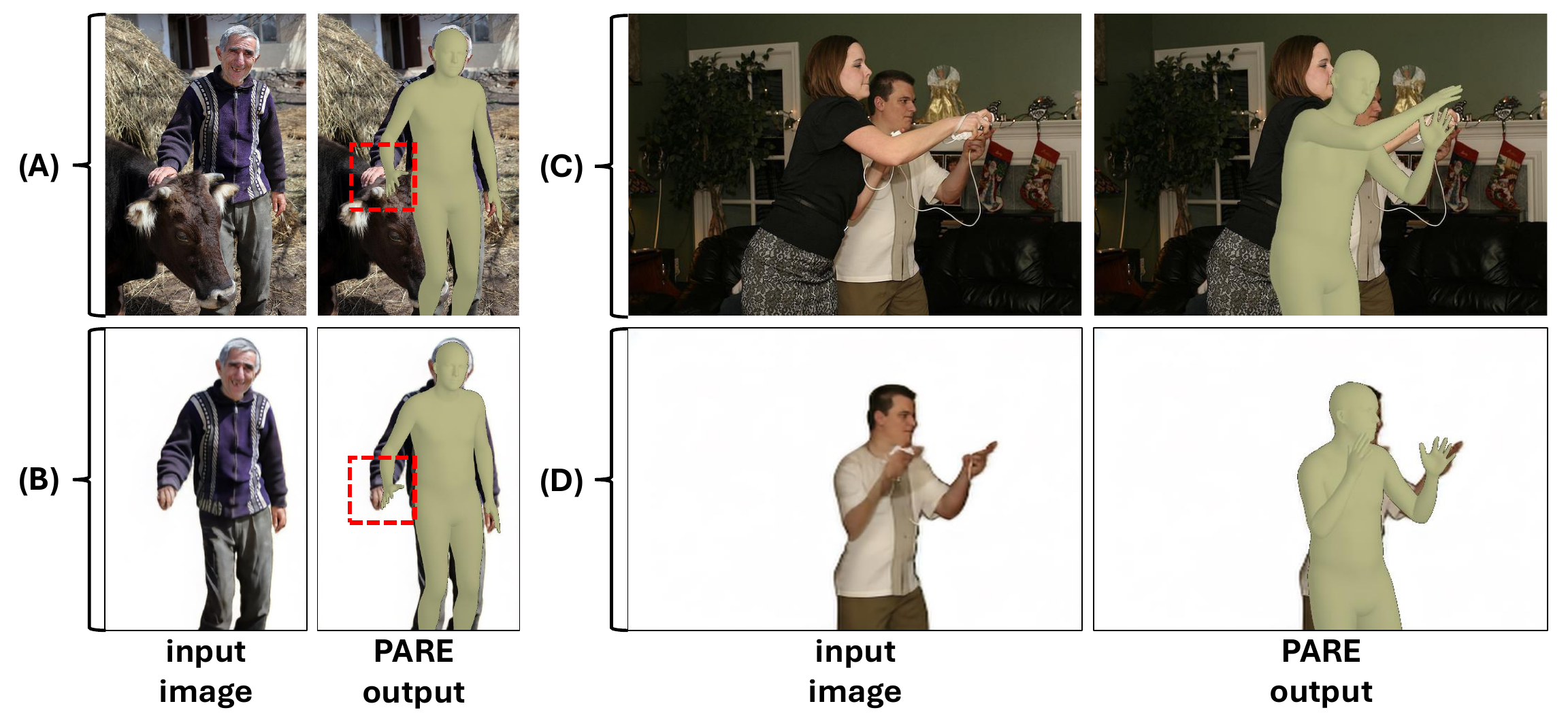}
    \vspace{-4mm}
    \caption{Qualitative results of occlusion-robust PSE~\cite{kocabas2021pare} with occluded and de-occluded inputs. (A,C): with occluded image, (B,D): with de-occluded image}
    \vspace{-4mm}
    \label{suppl_fig:m}
\end{figure}

\subsection{Effect on Occlusion-Robust Downstream Methods}

Sections~\ref{suppl_subsec:b9} and~\ref{suppl_subsec:c7} demonstrated that replacing the original occluded input image $I_\text{o}$ with the de-occluded image $I_\text{do}^*$ improves performance in various downstream tasks, using models that do not explicitly handle occlusion~\cite{saito2019pifu, 8765346, kanazawaHMR18}.

Building on this, our \emph{“de-occlusion first, then downstream task”} strategy supports reuse of existing frameworks without occlusion handling and also improves the performance of models explicitly designed to be occlusion-robust. To assess this, we evaluate PARE~\cite{kocabas2021pare}, an off-the-shelf method for pose and shape estimation (PSE) that incorporates occlusion-robust mechanisms. As illustrated in Fig.~\ref{suppl_fig:m}, PARE produces reasonable pose estimates in cases with moderate occlusion, such as the single-person example in (A), but its performance degrades in more challenging scenarios involving multiple persons or severe occlusion, as shown in (C). When provided with the de-occluded images generated by our method---shown in rows (B) and (D) of Fig.~\ref{suppl_fig:m}---PARE yields more complete and accurate reconstructions in both cases. These results indicate that de-occlusion can complement even occlusion-aware models by enhancing their robustness in complex scenes.

\begin{figure}
    \centering
    \includegraphics[width=0.5\linewidth]{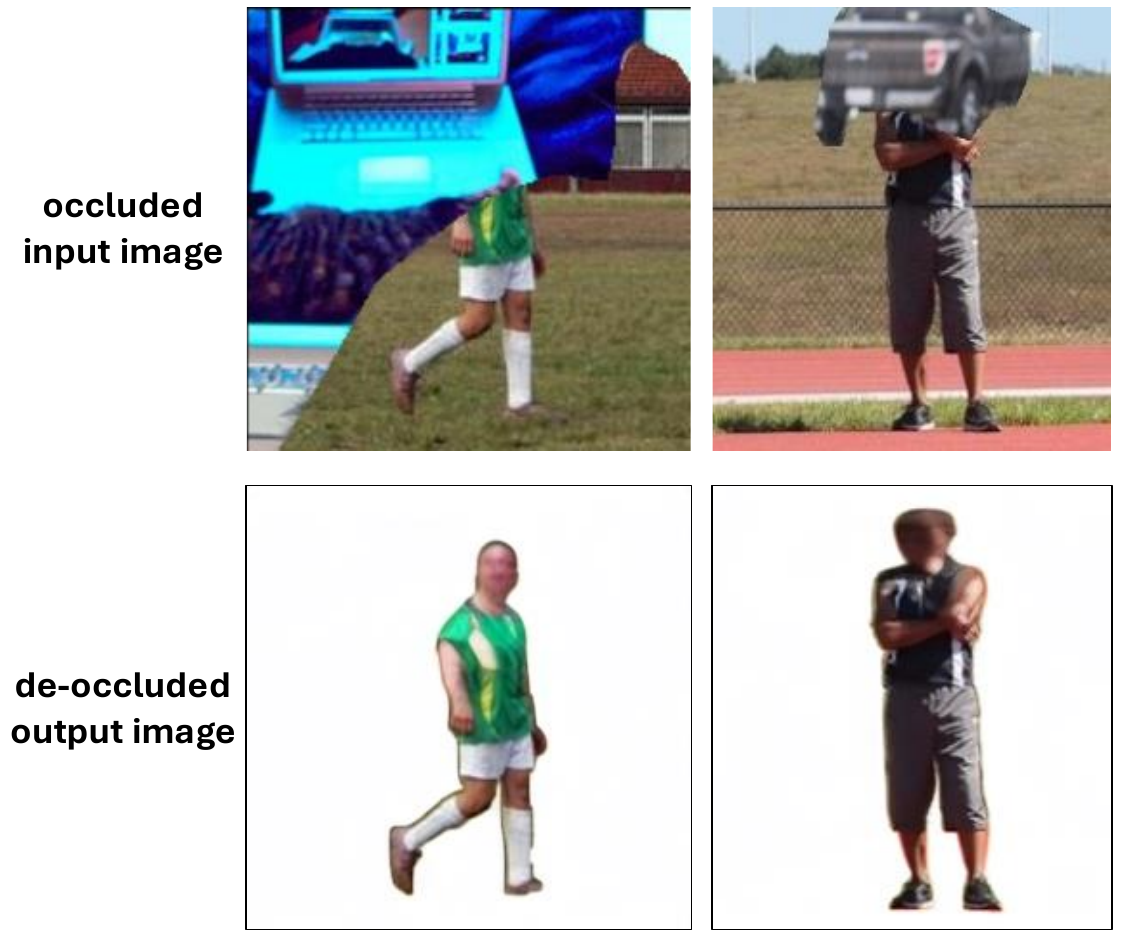}
    \vspace{-4mm}
    \caption{Failure cases under heavy facial occlusion}
    \vspace{-4mm}
    \label{suppl_fig:n}
\end{figure}

\subsection{Failure Cases and Limitations}

A limitation arises when the human face is heavily occluded, which makes it difficult to recover complex facial details. In our current RGB completion network, denoising is performed in a 32×32 latent space, where the face occupies a relatively small region. As a result, some fine-grained facial details may be lost during reconstruction through the VAE decoder. As shown in Fig.~\ref{suppl_fig:n}, the completed images generate a plausible global structure and appearance, but fail to recover fine-grained facial details. Incorporating face-specific priors could be a promising direction for future work.

\bibliographystyle{ACM-Reference-Format}
\balance
\bibliography{main}


\end{document}